\documentclass[10pt,twocolumn,letterpaper]{article}

\usepackage{cvpr}              

\usepackage[accsupp]{axessibility}
\usepackage{algorithm}   
\usepackage{algorithmicx}
\usepackage{algpseudocode}
\usepackage{amsmath}   
\usepackage{amssymb} 
\usepackage{multirow}
\definecolor{cvprblue}{rgb}{0.21,0.49,0.74}
\usepackage[table]{xcolor}
\usepackage{microtype}
\usepackage{marvosym}

\usepackage{listings}
\usepackage{xcolor}
\definecolor{hlt}{rgb}{0.9, 1.0, 0.9} 

\definecolor{codegreen}{rgb}{0, 0.6, 0}

\lstdefinestyle{paperstyle}{
    language=Python,
    basicstyle=\ttfamily\small,      
    keywordstyle=\color{codegreen}\bfseries, 
    stringstyle=\color{brown},       
    commentstyle=\color{gray},       
    breakatwhitespace=false,          
    breaklines=true,               
    captionpos=b,                    
    keepspaces=true,                 
    numbers=none,                 
    frame=none,                 
    xleftmargin=0pt,        
    xrightmargin=0pt,
    aboveskip=1em,             
    belowskip=1em,           
    showspaces=false,                
    showstringspaces=false,
    showtabs=false,                  
    tabsize=4
}

\usepackage[pagebackref,breaklinks,colorlinks,allcolors=cvprblue]{hyperref}
\definecolor{lightred}{RGB}{252, 221, 221}   
\definecolor{lightyellow}{RGB}{254, 252, 221}
\definecolor{lightgreen}{RGB}{221, 237, 213} 
\definecolor{lightblue}{RGB}{221, 236, 251} 

\title{MixFlow Training: Alleviating Exposure Bias
with Slowed Interpolation Mixture
}

\author{Hui Li$^{1,2,5}$~~~~~~~~~~Fu-Yun Wang$^3$~~~~~~~~~~Haoyuan Xia$^1$~~~~~~~~~~Jiayue Lyu$^2$~~~~~~~~~~Kaihui Cheng$^2$ \\
Siyu Zhu$^{1,2,5}$\textsuperscript{\Letter}~~~~~~~~~~Jingdong Wang$^{4}$\textsuperscript{\Letter}\\
$^1$Shanghai Innovation Institute~~~~~~$^2$Fudan University~~~~~~ 
$^3$The Chinese University of Hong Kong\\$^4$Baidu~~~~~~
$^5$Shanghai Academy of AI for Science\\
\url{https://mixflowgen.github.io/}}

\begin{document}
\maketitle
\renewcommand{\thefootnote}{\fnsymbol{footnote}}
\footnotetext[0]{\Letter~Corresponding authors}
\renewcommand{\thefootnote}{\arabic{footnote}}

\begin{abstract}
This paper studies the training-testing discrepancy 
(a.k.a. exposure bias) problem
for improving the diffusion models. 
During training,
the input of a prediction network
at 
one training timestep 
is 
the corresponding \emph{ground-truth} noisy data
that
is an interpolation
of the noise and the data,
and during testing,
the input is the \emph{generated} noisy data.
We present a novel training approach, named MixFlow, for improving the performance. 
Our approach is motivated by
the \emph{Slow Flow} phenomenon:
the ground-truth interpolation that is the nearest to the 
generated noisy data at a given sampling timestep
is observed to
correspond to a higher-noise timestep
(termed slowed timestep),
i.e.,
the corresponding ground-truth timestep is slower than the sampling timestep. 
MixFlow 
leverages the interpolations at
the slowed timesteps,
named~\emph{slowed interpolation mixture},
for post-training
the prediction network
for each training timestep.
Experiments over class-conditional image generation
(including SiT,
REPA,
and RAE)
and 
text-to-image generation
validate the effectiveness of our approach.
Our approach MixFlow over the RAE models achieve strong generation results on ImageNet: 
$1.43$ FID (without guidance)
and $1.10$ (with guidance) at $256 \times 256$,
and $1.55$ FID (without guidance)
and $1.10$ (with guidance) at $512 \times 512$.
\end{abstract}

\begin{figure}[t]
  \centering
  \footnotesize
    \centering
\subfloat[]{\includegraphics[width=0.47\linewidth]{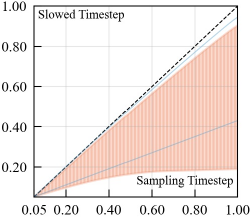}}
~~~~
\subfloat[]{\includegraphics[width=0.47\linewidth]{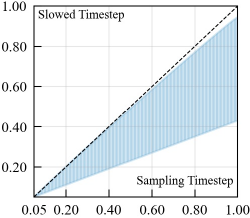}}
\vspace{-3mm}
  \caption{
  Illustrating (1) the~\emph{Slow Flow} phenomenon during the sampling process: the timestep (y-axis),
  corresponding to
  the ground truth noisy data
  that is the nearest to
  the generated noisy data at the sampling timestep $t$ (x-axis),
  is slower (with higher noise),
  \ie,
  the shading area is under the line $x=y$;
  and (2) the~\emph{effectiveness of MixFlow training}:  
  the range of slowed timesteps for (b) MixFlow training
  is smaller 
  and closer to the sampling steps
  than (a) standard training,
  indicating that MixFlow training
  effectively alleviates the 
  training-testing discrepancy.
  The boundary of the shading area in (b) is plotted as blue lines in (a).
  Note:
  x-axis - the sampling timestep at which the noisy data is generated;
  y-axis - the slowed timestep
  corresponding to
  the ground truth noisy data
  that is the nearest to
  the generated noisy data;
  shading area -  
  the range (the vertical line) of slowed timesteps 
  at each sampling step;
  noise corresponds to timestep $0$,
  and data corresponds to timestep $1$.
  The slowed timestep ranges 
  are obtained
  from $20,000$ training images in ImageNet~\cite{deng2009imagenet},
  $50$ sampling steps,
  and SiT-B~\cite{ma2024sit}.  
  Details on how
  to plot the figures
  are provided in Appendix~\ref{appendix:detailsforfigure1}.
}
\label{fig:slowflow}
\vspace{-3mm}
\end{figure}

\section{Introduction}
\label{sec:intro}

We study the training-testing discrepancy problem~\cite{ranzato2015sequence, schmidt2019generalization},
also known as exposure bias~\cite{huang2025selfforcing, ning2023input, ning2023epsilonscaling, li2023timeshift, lierror},
for diffusion and flow matching models.
During training, 
diffusion models learn a prediction network,
where the input to the prediction network 
at each training timestep is
the corresponding \emph{ground-truth} noisy data,
\ie,
an interpolation
of the noise and the data.
During testing,
the input to the prediction network
is the \emph{generated} noisy data.
The difference
of the inputs
to the prediction network 
for training and testing,
\ie,
the training-testing discrepancy,
is one of the reasons
leading to the prediction discrepancy
and accordingly
the problems of error accumulation and sampling drift.

There are two main lines of solutions to alleviating the discrepancy problem. 
One line is to modify the training procedure~\cite{ning2023input,huang2025selfforcing}.
For example,
Input Perturbation~\cite{ning2023input} conducts an input perturbation on the ground truth noisy data,
and 
self-forcing~\cite{huang2025selfforcing} uses the generated noisy data as the input. 
The other line is to modify the sampling process~\cite{ning2023epsilonscaling, li2023timeshift, Manifold, ren2024multi}.
For example,
Epsilon Scaling~\cite{ning2023epsilonscaling}
scales the predicted noise 
during sampling,
and Time-Shift Sampler~\cite{li2023timeshift}
shifts the sampling timestep for the next sampling iteration.
In this paper,
we are interested in the former line
and present a novel training procedure.

Our approach is motivated by the~\emph{Slow Flow} phenomenon about
the generated noisy data
and its nearest ground-truth noisy data.
\Cref{fig:slowflow} illustrates 
the~\emph{Slow Flow} phenomenon. 
The nearest ground-truth noisy data  to the generated noisy data at the sampling timestep $t$ 
is observed to corresponds to a higher-noise timestep, called~\emph{Slowed Timestep} $m_t$.
This intuitively means that
the generated noisy data is slower
than the ground truth noise data,
\ie, the higher-noise timestep slower than the sampling timestep
($m_t \leq t$).
In addition,
the range of the timestep difference is larger for a greater sampling timestep $t$,
meaning that
the slowed timestep at a greater timestep $t$
is possibly more different from 
the sampling timestep $t$.
\looseness=-1

In light of the observation,
we present a novel training method named MixFlow,
which leverages the ground-truth noisy data at slowed timesteps, called slowed interpolation,
for
training the prediction network.
The input noisy data
to the prediction network at one training timestep
is a mixture of slowed interpolations,
containing the interpolation
corresponding to the training timestep,
as well as the interpolations
at higher-noise timesteps.
In our implementation,
each slowed interpolation,
is uniformly sampled for any training timestep,
and the training timestep is sampled according to the probability $\operatorname{Beta}(2,1)$.
This implementation results in
that
each pair of slowed interpolation
and training timestep,
input to the prediction network, 
is uniformly sampled across 
all the pairs of slowed interpolation
and training timestep.
The implementation is very simple:
only a few lines of code are modified
for training with MixFlow.
For example,
only 5 lines are modified
for MixFlow + RAE~\cite{zheng2025rae}.

We demonstrate the effectiveness of MixFlow by post-training the generation models.
Our approach outperforms representative methods that are developed for alleviating the training-testing discrepancy.
Our approach consistently improves the performance
on class-conditional image generation
(such as SiT~\cite{ma2024sit},
REPA~\cite{yu2025repa},
and
the very-recently developed method RAE~\cite{zheng2025rae})
and text-to-image generation.
The MixFlow training approach over the RAE models achieve strong image generation results on ImageNet: 
$1.43$ FID (without guidance)
and $1.10$ (with guidance) at $256 \times 256$,
and $1.55$ FID (without guidance)
and $1.10$ (with guidance) at $512 \times 512$.

\section{Related Works}
\label{sec:related_work}

\noindent\textbf{Visual generation with diffusion and flow.}
Flow~\cite{lipman22lowmatching, papamakarios2021normalizing, liuflow, liu2023instaflow, xu2022poisson} and diffusion models~\cite{ho2020denoising,song2019generative, song2020score, rombach2022high, saharia2022photorealistic}
have been widely applied
for visual generation.
Many variants have been presented.
Prediction outputs include noise~\cite{ho2020denoising, rombach2022high}, data~\cite{kingma2021variational}, score~\cite{song2020score, song2019generative, song2025selective},
and velocity~\cite{lipman22lowmatching, liuflow}.
Noise scheduling includes:
variance preserving interpolation~\cite{song2020score},
and linear interpolation~\cite{lipman22lowmatching, liuflow}.
Sampling algorithms include:
DDIM~\cite{songdenoising}, ODE~\cite{lipman22lowmatching}, SDE~\cite{karras2022elucidating}, and so on.
Other studies include backbone~\cite{peebles2023DiT, ma2024sit, esser2024SD3, rombach2022high, yu2024image, kouzelis2025boosting},  guidance mechanisms~\cite{ho2021classifier,Karras2024autoguidance} and latent representation~\cite{yu2025repa, zheng2025rae, yao2025reconstruction, wang2025diffuse, wu2025representation, stoica2025contrastive, Leng_2025_ICCV, jiang2025no}.

\vspace{0.1cm}
\noindent\textbf{Alleviating exposure bias in diffusion and flow.}
The exposure bais issue is studied in diffusion models for visual generation.
The techniques include new training schemes, \eg, Input Perturbation~\cite{ning2023input}, and new inference schemes,
\eg, Epsilon Scaling~\cite{ning2023epsilonscaling},
Time-Shift Sampler~\cite{li2023timeshift} , manifold constraint-based sampler~\cite{Manifold} and multi-step denoising scheduled sampler~\cite{ren2024multi}.

Input Perturbation~\cite{ning2023input} trains the diffusion model
by 
conducting a Gaussian perturbation
on the ground truth noisy data 
to simulate the inference time prediction errors.
The performance is sensitive to the perturbation strength: too large or too small even lead to worse result.
Self Forcing~\cite{huang2025selfforcing},
a novel training paradigm for autoregressive video diffusion models,
uses previously-generated noisy data
as the input for training diffusion models.
As pointed out in Self Forcing~\cite{huang2025selfforcing},
it is suitable for few-step diffusion as
self forcing for
standard many-step diffusion models would be computationally prohibitive.

Epsilon Scaling~\cite{ning2023epsilonscaling} adjusts the sampling process
by scaling the noise prediction, mitigating the input mismatch between training and sampling.
Time-Shift Sampler~\cite{li2023timeshift}
modifies the sampling process
by finding a coupled timestep of the previously sampled data,
for the next sampling iteration.
The coupled timestep ideally 
corresponds to the nearest ground truth noisy data,
and is found heuristically 
in the sampling process. 
Differently, our approach addresses a related problem
by modifying the training procedure.

\section{Diffusion and Flow Matching}
Flow matching and diffusion models
adopt a stochastic process
that
transforms from the noise
to the data.
Usually,
the noise distribution is a Gaussian distribution:
$\mathbf{x}_0 \sim \mathcal{N}(0,1)$,
and the data distribution $\mathbf{x}_1 \sim q$
is represented by data samples.
The interpolation,
ground-truth noisy data,
in the stochastic process can be formulated as a combination
of the noise and the data:
\begin{align}
    \mathbf{x}_t = \alpha_t \mathbf{x}_1
    + \beta_t \mathbf{x}_0,
    \label{eqn:standinterpolation}
\end{align}
where $\alpha_t$ increases 
when $t$ increases
and $\beta_t$ decreases
when $t$ increases.
Flow matching
adopts the following setting:
\begin{align}
    \alpha_t = t,~~\beta_t = 1 - t.
    \label{eqn:flowmatching}
\end{align}
Variance-preserving diffusion models
use
$\alpha_t^2 + \beta_2^2 = 1$,
\eg,
generalized variance-preserving diffusion models~\cite{ma2024sit}
set:
\begin{align}
    \alpha_t=\sin(\frac{1}{2}\pi t),~~
    \beta_t=\cos(\frac{1}{2}\pi t).
    \label{eqn:gvp}
\end{align}

The model prediction output could be the noise, the data, the score, or the velocity. The loss function, using the velocity as the prediction target,
is as follows,
\begin{align}
    \mathbb{E}_{t, \mathbf{x}_0,\mathbf{x}_1}[\|\mathbf{u}_{\boldsymbol{\theta}}(\mathbf{x}_t, t) - \mathbf{u}^*(\mathbf{x}_t, t)\|_2^2],
    \label{eqn:standardloss}
\end{align}
where $\mathbf{u}^*(\mathbf{x}_t, t)$
is the ground truth velocity
for the interpolation $\mathbf{x}_t$
at the training timestep $t$.

The typical sampling
algorithms
include: reverse-time stochastic differential equation (SDE) sampler, $\mathrm{d}\mathbf{x}_t
= \mathbf{u}_{\boldsymbol{\theta}}(\mathbf{x}_t,t)\mathrm{d}t
- \frac{1}{2}\bar{w}_t\mathbf{s}(\mathbf{x}_t,t)\mathrm{d}t + \sqrt{\bar{w}_t}
\mathrm{d}\bar{\mathbf{w}}_t$,
where $\bar{\mathbf{w}}_t$
is 
a reverse-time Wiener process,
$\bar{w}_t > 0$ is an arbitrary time-dependent diffusion coefficient,
$\mathbf{s}(\mathbf{x}_t,t)$ is the score;
a probability flow ordinary differential equation (ODE) sampler,
$\mathrm{d}\mathbf{x}_t
= \mathbf{u}_{\boldsymbol{\theta}}(\mathbf{x}_t,t)\mathrm{d}t$.
The ODE can be numerically solved using the first-order Euler's method:
\begin{align}
\mathbf{x}_{i+1} = \mathbf{x}_i + \Delta t \mathbf{u}_{\boldsymbol{\theta}}(\mathbf{x}_i, t_i), 
\end{align} or the second-order Heun's method:
\begin{equation}
\left\{
\begin{aligned}
\tilde{\mathbf{x}}_{i+1} &= \mathbf{x}_i + \Delta t  \mathbf{u}_{\boldsymbol{\theta}}(\mathbf{x}_i, t_i) \\
\mathbf{x}_{i+1} &= \mathbf{x}_i + \frac{\Delta t}{2} \left[ \mathbf{u}_{\boldsymbol{\theta}}(\mathbf{x}_i, t_i) + \mathbf{u}_{\boldsymbol{\theta}}(\tilde{\mathbf{x}}_{i+1}, t_{i+1}) \right],
\end{aligned}
\right.
\end{equation}
where $i$ is timestep index, and $\Delta t = t_{i+1} - t_i$ is step size.

\section{MixFlow}
\label{sec:method}
The proposed MixFlow approach
is inspired
by the~\emph{Slow Flow} phenomenon
shown in \Cref{fig:slowflow}.
MixFlow includes several key points:
slowed interpolation mixture
for leveraging the ground truth noisy data
from slowed timesteps,
\ie, high-noise timesteps,
for training the prediction network
at each training timestep;
loss function formulated with slowed interpolation mixture;
as well as
slowed timestep 
and training timestep sampling.

\vspace{1mm}
\noindent\textbf{Slowed interpolation mixture.}
Standard diffusion and flow matching models adopt a \emph{single}
interpolation\footnote{In this paper,
the two terms,
interpolation 
and ground truth noisy data
are about the same meaning
without special description.}  $\mathbf{x}_t$ at the training timestep $t$ as input for training the prediction network, \eg, the velocity prediction network $\mathbf{u}_{\boldsymbol{\theta}}(\mathbf{x}, t)$ in our study.
Given the data $\mathbf{x}_1$ and the noise $\mathbf{x}_0$,
the interpolation is computed as 
\Cref{eqn:standinterpolation},
and the interpolation coefficients $\alpha_t$
, $\beta_t$ only correspond to the training timestep $t$.
\looseness=-1

Our approach, instead,
uses an infinite mixture of interpolations
as input for training
$\mathbf{u}_{\boldsymbol{\theta}}(\mathbf{x}, t)$ at training timestep~$t$:
\begin{align}
    \mathcal{X}_{t} = \{
    \mathbf{x}_{m_t} | 
    \mathbf{x}_{m_t} = \beta_{m_t} \mathbf{x}_0
    + \alpha_{m_t} \mathbf{x}_1,
    m_t \in \mathcal{M}_t\},
\end{align}
where $\mathcal{M}_t$ is the range of timesteps, relying on timestep $t$.

The timestep range $\mathcal{M}_t$
is formed 
according to the~\emph{Slow Flow} phenomenon shown in \Cref{fig:slowflow}:
The timestep corresponding to
the ground truth interpolation that is nearest
to the generated noisy data at 
the sampling timestep $t$
is observed to be smaller than $t$, \ie,
higher-noise timestep
which we call~\emph{slowed timestep};
The slowed timestep range
at greater sampling timestep $t$
is larger,
\ie,
the slowed timestep difference between the generated noisy data
and the ground truth noisy data
could be larger at greater sampling timestep.

We choose~\emph{slowed interpolation mixture}: a set of interpolations
from a range of slowed timesteps:
\begin{align}
\mathcal{M}_t 
= [(1-\gamma)t, t]. 
\end{align}
The mixture range size is: $t - (1-\gamma)t = \gamma t$,
which is linear 
with respect to the training timestep $t$.
$\gamma$ can be empirically selected or simply set as $1$.

\begin{figure}[t]
  \centering
  \footnotesize
  \subfloat[]{\includegraphics[scale =1]{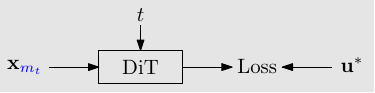}}~~~~~~
  \subfloat[]{\includegraphics[scale =1]{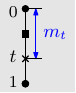} }
   \caption{
   Illustrating
   MixFlow
   training.
   (a) MixFlow training.
   At the training timestep $t$,
   the input noisy data is the interpolation
   $\mathbf{x}_{m_t}$ at the slowed timestep $m_t$.
   It is different from standard training:
   the input noisy data is the interpolation
   $\mathbf{x}_t$ at the timestep $t$.
   (b) Slowed timestep.
   $m_t$ is a timestep lying in the range $[(1-\gamma)t, t]$.
   The illustration in (b) is for $\gamma = 1$.   
}   \label{fig:standandmixflowtraining}
\end{figure}

\vspace{0.1cm}
\noindent\textbf{Loss function.}
Our approach with slowed interpolation mixture
optimizes the following loss function,
\begin{align}
    \mathbb{E}_{t, \mathbf{x}_0,\mathbf{x}_1,m_t}[\|\mathbf{u}_{\boldsymbol{\theta}}(\mathbf{x}_{m_t}, t) - \mathbf{u}^*(\mathbf{x}_{m_t}, m_t)\|_2^2].
    \label{eqn:mixflowloss}
\end{align}
In comparison
to the standard loss function in \Cref{eqn:standardloss},
an extra variable,
slowed timestep $m_t$,
is introduced.
The loss is
not only about 
the interpolation
at the training timestep $t$,
but also
the slowed interpolation $\mathbf{x}_{m_t}$
at the higher-noise (slowed) timestep $m_t$.

The input noisy data to the prediction network $\mathbf{u}_{\boldsymbol{\theta}}$
is the interpolation $\mathbf{x}_{m_t}$
at a slowed timestep $m_t$
instead of 
the interpolation
$\mathbf{x}_t$
at the training timestep $t$
in standard training
(\Cref{eqn:standardloss}).
The input timestep is still the training timestep $t$
(other than 
the slowed timestep $m_t$).
It should be noted that 
the sampling process is the same
with that for standard training
and there is no need to compute
slowed timestep $m_t$.
\Cref{fig:standandmixflowtraining}
illustrates
of MixFlow training.

\begin{figure*}[t]
  \centering
  \footnotesize 
  \captionsetup[subfloat]{
labelformat=empty} 
\vspace{-2mm}

  \subfloat[$t=0.2$]{
\includegraphics[width=0.19\linewidth]{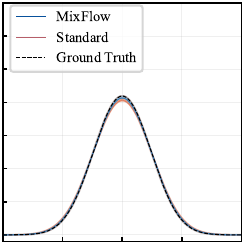}}
~
\subfloat[$t=0.4$]{
\includegraphics[width=0.19\linewidth]{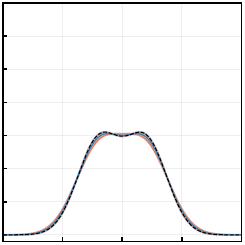}
}
~
\subfloat[$t=0.6$]{\includegraphics[width=0.19\linewidth]{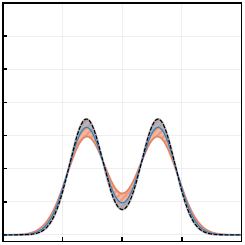}}
~
\subfloat[$t=0.8$]{
\includegraphics[width=0.19\linewidth]{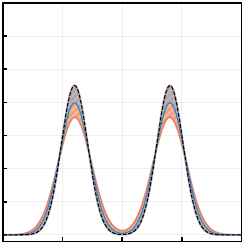}}
~
\subfloat[$t=1.0$]{    \includegraphics[width=0.19\linewidth]{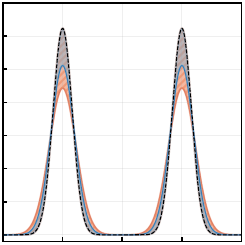}}
  \caption{A toy example 
  illustrating the advantage
  of the MixFlow training
  over the standard training.
  The distribution
  from the model with the MixFlow training 
  fits better the ground truth distribution than
  the standard training.
  Details
  are provided in Appendix~\ref{appendix:detailsforfigure3}.}
  \label{fig:toyexample}
\vspace{-2mm}

\end{figure*}

\vspace{0.1cm}
\noindent\textbf{Training.}
The MixFlow loss function in \Cref{eqn:mixflowloss}
introduces the extra variable $m_t$,
the slowed timestep,
which we need to handle for MixFlow training. 
We sample $m_t$
at the training timestep $t$
from a simple distribution
$p(m_t | t)$,
a uniform distribution:
\begin{align}
m_t  \sim \mathcal{U}
[(1-\gamma)t, t].
\end{align}

The mixture range size
of the slowed timesteps $m_t$ at the training timestep $t$
is $\gamma t$.
The conditional probability is
thus $p(m_t | t) = \frac{1}{\gamma t}$
for $m_t \in [(1-\gamma)t, t]$.
We expect that
all the pairs $(\mathbf{x}_{m_t}, t)$,
the inputs to the prediction network $\mathbf{u}_{\boldsymbol{\theta}}$,
are evenly sampled:
$p(m_t, t) = p(m_{t'}, t')$.
Considering that
$p(m_t, t) = p(m_t | t) p(t)$,
we have $p(t) \propto t$.
Since $\int_0^1 p(t) = 1$
and $t \in [0, 1]$,
$p(t) = 2t$,
meaning that $p(t)$ is a Beta distribution
\looseness=-1
\begin{align}
t \sim 
\operatorname{Beta}(2,1).
\end{align}
The algorithm
is given in Algorithm~\ref{alg:mixflow}.
The key differences from 
standard training
lie in 
Line $5$ - Line $8$.

\begin{algorithm}[t]
\caption{MixFlow Training}
\label{alg:mixflow}
\begin{algorithmic}[1]
\State \textbf{Input:}  
dataset $\mathcal{D}$, 
initial model parameter $\boldsymbol{\theta}$,
mixture range coefficient $\gamma$, 
learning rate $\eta$
 
\Repeat
\State Sample $\mathbf{x}_1 \sim \mathcal{D}$
\State Sample $\mathbf{x}_0 \sim \mathcal{N}( \mathbf{0}, \mathbf{I})$
\State Sample $t \sim \operatorname{Beta}(2,1)$
\State Sample  $m_t \sim \mathcal{U}[(1-\gamma)t, t]$
\State $\mathbf{x}_{m_t} \leftarrow 
\beta_{m_t}\mathbf{x}_0 + \alpha_{m_t}  \mathbf{x}_1$
\State $\mathcal{L} \leftarrow \| \mathbf{u}_{\boldsymbol{\theta}}(\mathbf{x}_{m_t}, t) - \mathbf{u}^*\|_2^2$
\State $\boldsymbol{\theta} \leftarrow \boldsymbol{\theta}- \eta \nabla_{\boldsymbol{\theta}} \mathcal{L}$
\Until convergence
\end{algorithmic}
\end{algorithm}

\vspace{1mm}
\noindent\textbf{Toy example.}
We use a $1$D toy example
to show that MixFlow training can generate samples
that fit the data distribution better than standard training.
The distribution for data $x_1$
is a mixture of two Gaussians: $p(x_1) = 0.5 \mathcal{N}(x_1; -2, 0.1^2) + 0.5 \mathcal{N}(x_1; 2, 0.1^2)$.
The distribution for noise $x_0$
is a Gaussian: $p(x_0) = \mathcal{N}(x_0; 0, 1)$.

\Cref{fig:toyexample}
illustrates the distributions
of generated noisy data
at five sampling timesteps 
for MixFlow training,
standard training,
and ground truth.
The observations include:
MixFlow learns
a better distribution than
standard training;
The discrepancy between 
the learned distribution and 
the ground truth distribution 
becomes larger at greater timesteps.
The observations
are consistent to
\Cref{fig:slowflow}:
MixFlow trains the model better;
The timestep difference range, between
the nearest ground truth data
and the generated noisy data,
becomes larger for 
larger sampling timestep.

\section{Experiments}
\label{sec:experiments}

\subsection{Ablation Studies}
\label{sec:ablation}
We study key components in MixFlow
over the SiT-B model~\cite{ma2024sit}:
sampling distributions
for training timestep $t$ and slowed timestep $m_t$,
and
mixture range coefficient $\gamma$.
We implement 
the MixFlow training algorithm
by 
modifying the official SiT implementation,
and 
post-train the pretrained
model
in $500K$ steps on ImageNet~\cite{deng2009imagenet}.
We use 
the same training hyperparameters,
the same evaluation setting,
the same ODE sampling process
(second-order Heun sampler, 250 sampling steps)
and the same classifier-free guidance scale ($1.5$) as
the original setting in SiT~\cite{ma2024sit}.
The metrics include
gFID~\cite{heusel2017gans}, sFID~\cite{sFID}, IS~\cite{salimans2016improved}, Precision and Recall~\cite{kynkaanniemi2019improved}
for class-conditional generation,
and GenEval~\cite{ghosh2023geneval}, DPG-Bench~\cite{hu2024ella} and T2I-CompBench~\cite{huang2023t2icompbench}
for text-to-image generation.

\vspace{1mm}
\noindent\textbf{Sampling distributions for training timestep $t$ 
and slowed timestep $m_t$.}
MixFlow samples the training timestep $t$ 
from a Beta distribution $\operatorname{Beta}(2,1)$,
and samples the slowed timestep $m_t$
from a uniform distribution 
$\mathcal{U}[(1-\gamma)t, t]$.
In this study, we
set mixture range coefficient $\gamma$ as $1$,
\ie, $m_t \sim \mathcal{U}[0, t]$,
and the choice of mixture range coefficient $\gamma$
will be studied later.
We study alternative sampling distributions:
$t \sim \mathcal{U}[0,1]$
and $m_t \sim \mathcal{U}[0,1]$.
\Cref{tab:sampledistributions}
shows the study results. 
\looseness=-1

Sampling the timestep $m_t$ from $\mathcal{U}[0,t]$,
\ie,
only higher-noise timesteps can be sampled
for interpolation mixture,
yields better performance than sampling $m_t$ from $\mathcal{U}[0,1]$,
\ie,
both higher-noise and lower-noise timesteps 
can be sampled. 
The result for sampling $m_t$ from $\mathcal{U}[0,1]$,
$18.27 / 5.07$ and $18.25 / 5.06$
is even worse than
the result of 
the baseline model,
$17.97 / 4.46$.
This indicates that
the inclusion of the interpolations
at lower-noise timesteps
harms the training,
\ie,
harming in solving the~\emph{Slow Flow} problem.

In the case that $m_t \sim \mathcal{U}[0, t]$,
sampling training timestep $t$
from $\mathcal{U}[0, 1]$
performs better than the baseline:
$16.58 / 4.25$ vs $17.97 / 4.46$
but worse than 
sampling $t$ from 
$\operatorname{Beta}(2,1)$:
$16.58 / 4.25$ vs $15.64 / 3.93$.
This implies that
evenly sampling the pair $(m_t, t)$
benefits the training,
which is consistent
to the analysis
in \Cref{sec:method}.

\begin{table}[]
    \centering
    \footnotesize   
    \setlength{\tabcolsep}{16pt} 
    \caption{Studies of
    sampling distributions 
    for $t$ and $m_t$ in 
    the MixFlow training.
    The metric scores
    are for generation
    without guidance /
    generation with guidance.
    }
\vspace{-2mm}
    \label{tab:sampledistributions}
    \begin{tabular}{c|c|c}
    \toprule
         & $t \sim \mathcal{U}[0,1]$ & $t \sim \operatorname{Beta}(2,1)$  \\
         \midrule
         $m_t \sim \mathcal{U}[0,1]$ & 18.27 / 5.07  & 18.25 / 5.06\\
         $m_t \sim \mathcal{U}[0,t]$ & 16.57 / 4.25 & \cellcolor{gray!20} 15.64 / 3.93\\
    \bottomrule
    \end{tabular}
    \vspace{-2mm}
\end{table}

\begin{figure}[t]
    \centering
  \footnotesize    \includegraphics
  [scale=0.95]{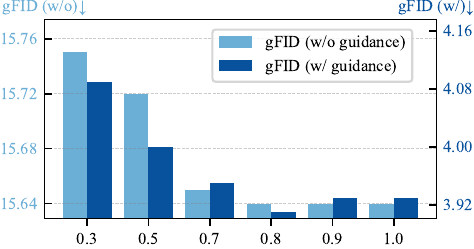}
\vspace{-3mm}
    \caption{Studies
    of mixture range coefficient $\gamma$ for sampling the slowed timestep: $m_t \sim \mathcal{U}[(1-\gamma)t, t]$.
    }
\label{fig:gamma}
\vspace{-3mm}
\end{figure}

\vspace{1mm}
\noindent\textbf{Mixture range coefficient $\gamma$ in the distribution
$\mathcal{U}[(1-\gamma)t, t]$
for sampling $m_t$.}
We consider seven values:
$\{0.3,
0.5,
0.7,
0.8,
0.9,
1.0
\}$. 
We do not show the results
for $\gamma = 0$,
which is slightly worse than baseline 
($18.00 / 4.48$ vs $17.97 / 4.46$).
The results are given in
\Cref{fig:gamma}.
$\gamma = 0.7,0.8,0.9,1.0$ performs better and similarly.
$\gamma = 0.8$ performs slightly better:
the gFID scores without guidance
for $\gamma = 0.7, 0.8,0.9,1.0$
are $15.65, 15.64, 15.64, 15.64$
and 
the scores with guidance 
are $3.95, 3.91, 3.93, 3.93$.
The slightly-better performance 
might come from
that the interpolations
from the over-slowed timesteps
\eg, $m_t \in [0, 0.2 t]$,
are not helpful as
for 
the model from standard training
at the sampling step $t$ it is unlikely to be delayed to $[0, 0.2t]$,
which is consistent to the observation 
as shown in \Cref{fig:slowflow}.
We choose $\gamma = 0.8$
as the choice
if not specified
for the following experiments.
\looseness=-1

\vspace{1mm}
\noindent\textbf{MixFlow training vs standard training.}
One may have a question:
does the performance get improved if the same number of standard post-training steps are conducted.
The results shown in \Cref{tab:mixflowvsstandard} suggest 
that
standard post-training almost does not affect the performance,
and the gain from MixFlow post-training
does not come from more training steps,
but comes from
the MixFlow training scheme.

\begin{table}[]
    \centering
    \footnotesize    
    \setlength{\tabcolsep}{13pt} 
    \caption{Comparison of
    standard
    and MixFlow post-training.}
\vspace{-2mm}
    \begin{tabular}{l|c|c}
    \toprule
          & w/o guidance &  w/ guidance  \\
         \midrule
         No post-training & 17.97 & 4.46 \\
         Standard post-training & 17.96 & 4.46 \\
         \rowcolor{gray!20}
         MixFlow post-training & 15.64 & 3.91  \\
         \bottomrule
    \end{tabular}
    \label{tab:mixflowvsstandard}
\vspace{-6mm}
\end{table}

\subsection{Improvement
by
MixFlow Training}
\label{sec:improvementbymixflow}
We validate
the effectiveness
of 
the MixFlow training 
by improvements over class-conditional generation
and text-to-image generation models
and comparison to methods
to exposure bias alleviation.

\noindent\textbf{Improvement over various class-conditional generation models.}
We post-train the prior SOTA 
class-conditional generation models
that are from the officially-released models at GitHub,
including SiT with linear and variance-preserving interpolations~\cite{ma2024sit},
REPA~\cite{yu2025repa},
and RAE~\cite{zheng2025rae}.
We follow the training settings,
including learning rate, batch size,
which are used in the official implementation for each model. 
Without special description,
the evaluation settings are the same:
same sampling scheme (second-order Heun sampler),
same guidance scale ($1.5$),
and 
sample sampling steps ($250$).

\begin{table*}[t]
\footnotesize
\setlength{\tabcolsep}{1pt} 
\caption{\textbf{The performance of MixFlow 
over two model scales: SiT-B and SiT-XL}
for class-conditional generation on ImageNet.}
\vspace{-2mm}
\begin{tabular*}{\textwidth}{@{} l  @{\extracolsep{\fill}}  r r r r r r r r r r@{}}
\toprule
  \multirow{2}{*}{\textbf{ Method }} &  \multicolumn{5}{c}{\textbf{Generation  w/o guidance}} & \multicolumn{5}{c}{\textbf{Generation  w/ guidance}} \\
 \cmidrule(lr){2-6} \cmidrule(lr){7-11}
    & gFID $\downarrow$ & sFID$\downarrow$ & IS $\uparrow$ & Pre. $\uparrow$ & Rec. $\uparrow$ & gFID $\downarrow$ & sFID $\downarrow$ & IS $\uparrow$ & Pre. $\uparrow$ & Rec. $\uparrow$ \\
\midrule
\multicolumn{11}{l}{\emph{ImageNet $256 \times 256$}}\\
\arrayrulecolor{black!30}\midrule
SiT-B/2   & 17.97&6.43& 79.4 &0.62&\textbf{0.66} &4.46 &4.93 & 182.4 &	0.78&\textbf{0.57 } \\
+ MixFlow  & \textbf{15.64} & \textbf{5.21} & \textbf{92.0} & \textbf{0.63} &\textbf{ 0.66} & \textbf{3.91} & \textbf{4.76} & \textbf{201.7} & \textbf{0.79} &0.56 \\
\arrayrulecolor{black!30}\midrule
SiT-XL/2   &  9.35 & 6.38 & 126.1 & 0.67 &\textbf{ 0.68} & 2.15 & 4.60 & 258.1 & \textbf{0.81} & 0.60\\
+ MixFlow   &  \textbf{7.87} & \textbf{4.91} & \textbf{140.3} & \textbf{0.68} & 0.67 & \textbf{1.99} & \textbf{4.34} & \textbf{272.0} & 0.80 & \textbf{0.61}  \\
\arrayrulecolor{black}\midrule
\multicolumn{11}{l}{\emph{ImageNet $512 \times 512$}}
\\
\arrayrulecolor{black!30}\midrule
SiT-XL/2  &  9.72  & 7.50  & 119.1 & 0.78 & \textbf{0.64} & 2.71 & 4.25 & 239.1 & 0.83 & \textbf{0.54}\\
+ MixFlow & \textbf{7.99} &\textbf{ 6.53 }& \textbf{131.0 }& \textbf{0.79 }&\textbf{ 0.64} & \textbf{2.53} & \textbf{4.21} & \textbf{246.2} & \textbf{0.84} & \textbf{0.54}\\
\arrayrulecolor{black}\bottomrule
\end{tabular*}
\label{tab:sitflowmatchingODEsolver}
\vspace{-4mm}
\end{table*}

\begin{table}[]
    \centering
    \footnotesize   
    \setlength{\tabcolsep}{7.2pt} 
    \caption{\textbf{The performance of MixFlow 
over various sampling steps}: 250, 50 and 20.
The gain is more significant for fewer steps.}
\vspace{-2mm}
    \begin{tabular}{lllllll}
    \toprule
          \multirow{2}{*}{\textbf{ Method }}& \multicolumn{3}{c}{\textbf{gFID  w/o guidance}} & \multicolumn{3}{c}{\textbf{gFID  w/ guidance}} \\
           \cmidrule(lr){2-4} \cmidrule(lr){5-7}
         & 250 & 50 & 20 & 250 & 50 & 20  \\
         \midrule
         \multicolumn{7}{l}{\emph{ImageNet $256 \times 256$}}\\
         \arrayrulecolor{black!30}\midrule
         SiT-B/2  & 17.97 & 18.10 & 19.05  & 4.46 & 4.57 & 4.88  \\
         + MixFlow  & 15.64 & 15.76 & 16.47 & 3.91 & 3.93 & 4.10  \\
         \textit{Gain} & \textbf{2.33} & \textbf{2.34} & \textbf{2.58} & \textbf{0.55} & \textbf{0.64} & \textbf{0.78} \\
    \arrayrulecolor{black!30}\midrule
         SiT-XL/2  &  9.35 & 9.73 & 10.67 & 2.15 & 2.17 & 2.28 \\
         + MixFlow & 7.87 & 8.00 & 8.71 & 1.99 & 2.01 & 2.09 \\
         \textit{Gain} & \textbf{1.48} & \textbf{1.73} & \textbf{1.96} & \textbf{0.16} & \textbf{0.16} & \textbf{0.19}\\
         \arrayrulecolor{black}\midrule
         \multicolumn{7}{l}{\emph{ImageNet $512 \times 512$}}\\
         \arrayrulecolor{black!30}\midrule
         SiT-XL/2 & 9.72 & 11.12 & 13.50 & 2.71 & 2.83 & 3.12\\
         + MixFlow &7.99 & 9.01 & 10.82 & 2.53 & 2.59 & 2.84 \\
         \textit{Gain} & \textbf{1.73} & \textbf{2.11} & \textbf{2.68} & \textbf{0.18} & \textbf{0.24} & \textbf{0.28}  \\
         \arrayrulecolor{black}\bottomrule
    \end{tabular}
    \label{tab:NFE_gain}
\vspace{-5mm}
\end{table}

\begin{table*}[!th]
\footnotesize
\setlength{\tabcolsep}{1pt} 
\caption{\textbf{The effectiveness of MixFlow
on SiT with generalized variance-preserving interpolations,
REPA and RAE.}} 
\label{tab:GVPREPARAE}
\vspace{-2mm}
\begin{tabular*}{\textwidth}{@{} l @{\extracolsep{\fill}}  c c c c c  c c c c c@{}}
\toprule
\multirow{2}{*}{\textbf{ Method }}&   \multicolumn{5}{c}{\textbf{Generation  w/o guidance}} & \multicolumn{5}{c}{\textbf{Generation  w/ guidance}} \\
 \cmidrule(lr){2-6} \cmidrule(lr){7-11}
   & gFID $\downarrow$ & sFID$\downarrow$ & IS $\uparrow$ & Pre. $\uparrow$ & Rec. $\uparrow$ & gFID $\downarrow$ & sFID $\downarrow$ & IS $\uparrow$ & Pre. $\uparrow$ & Rec. $\uparrow$ \\
\midrule
\multicolumn{11}{@{}l}{\textit{Diffusion}} \\
\arrayrulecolor{black!30}\midrule
SiT-B/2-GVP & 18.01 &	6.38&	79.39 &	0.61&	\textbf{0.66} & 4.55 &	4.91	&181.60 &	0.78	&\textbf{0.56}\\
+ MixFlow  & \textbf{15.59} & \textbf{6.31} & \textbf{90.52} & \textbf{0.63} & 0.64 & \textbf{4.35} & \textbf{4.88} & \textbf{199.94} & \textbf{0.79} & 0.54 \\
\arrayrulecolor{black}\midrule
\multicolumn{11}{@{}l}{\textit{REPA}}\\ \arrayrulecolor{black!30}\midrule
REPA-XL/2  & 6.90 & 6.03 & 148.9 & 0.68 & \textbf{0.69} & 1.65 & 4.63 & 278.3 & 0.77 & \textbf{0.65}\\
+ MixFlow  &  \textbf{6.28} & \textbf{5.06} & \textbf{159.8} & \textbf{0.69} &\textbf{ 0.69} & \textbf{1.59} & \textbf{4.34} & \textbf{290.5} & \textbf{0.78} & \textbf{0.65} \\
\arrayrulecolor{black}\midrule
\multicolumn{11}{@{}l}{\textit{RAE}}\\
\arrayrulecolor{black!30}\midrule
RAE-XL & 1.51 & 5.31 & \textbf{242.9} & 0.79 & 0.63 & 1.13 & 4.74 & \textbf{262.6} & \textbf{0.78} & \textbf{0.67} \\
+ MixFlow & \textbf{1.43} & \textbf{4.81} & 239.8 & \textbf{0.80} & \textbf{0.64} & \textbf{1.10} & \textbf{4.40} & 259.7 & 	\textbf{0.78}	& \textbf{0.67} \\
\arrayrulecolor{black}\bottomrule
\end{tabular*}
\vspace{-2mm}
\end{table*}

\noindent\emph{SiT.} 
We report the results on two model sizes:
SiT-B (130M)
and SiT-XL (675M).
The results on ImageNet $256 \times 256$
are reported in \Cref{tab:sitflowmatchingODEsolver}.
On can see that 
the overall performance of MixFlow is better across the two model sizes 
for generation without and with guidance.
The gFID scores for
generation with guidance
are improved
from $4.46$ to $3.91$
for SiT-B/2,
and from $2.15$
to $1.99$ for SiT-XL/2.

We also report the results on ImageNet $512 \times 512$
over the SiT-XL model.
The observation is consistent. 
MixFlow improves the performance, 
enhancing the gFID score without guidance 
from $9.72$ to $7.99$ and the gFID score with guidance from $2.71$ to $2.53$.

\Cref{tab:NFE_gain}
shows the results
for other sampling steps,
$\{250, 50, 20\}$.
One observation is that
the gains from MixFlow
become greater
for fewer sampling steps.
The reason is that
the~\emph{Slow Flow} issue
is more pronounced
for fewer sampling steps,
where the sampler approximation is coarser
and the training and testing difference is larger.

\noindent\emph{SiT-diffusion.}
We report the results of a diffusion model,
SiT with generalized variance-preserving interpolations
(\Cref{eqn:gvp}).
We consider another target choice
for the ground-truth velocity $\mathbf{u}^*$:
$\mathbf{u}^*(\mathbf{x}_{t}, t)$.
We find that the new choice 
is able to accelerate the training process:
the new choice can reach the same performance with $500K$ training steps
while 
the old choice needs 
$2500$K training.
We speculate that the new choice helps optimization
as in the GVP case  $\mathbf{u}^*(\mathbf{x}_{m_t}, m_t)$
varies along the path.
\Cref{tab:GVPREPARAE}
shows the results.
The observations are consistent
to flow matching
that uses the linear interpolation~in \Cref{eqn:flowmatching}:
the overall performance gets improved
for both generation without guidance
and with guidance.
The gFID scores
are improved from
$18.01$ to $15.59$ for generation
without guidance,
and from $4.55$ to $4.35$
for generation with guidance.
This experiment
demonstrates
that MixFlow is applicable
to diffusion models with variance preserving interpolations
besides linear interpolations.
\looseness=-1

\noindent\emph{REPA.}
We demonstrate the effectiveness
of MixFlow
over the models
that are trained
with advanced schemes.
We post-train the officially-released model trained
with REPA.
The post-training loss function
consists of the alignment loss used in REPA~\cite{yu2025repa}
as well as the MixFlow loss.
The evaluation schemes are the same as REPA~\cite{yu2025repa}, such as guidance scale ($1.8$) and guidance interval ($[0,0.7]$).

The results are given in \Cref{tab:GVPREPARAE}.
Almost all the scores
get improved except
that the recall scores  are the same.
The gFID scores
are improved:
$6.90 \rightarrow 6.28$
for generation without guidance,
and $1.65 \rightarrow 1.59$
for generation with guidance.
The performance improvement
indicates that
MixFlow is compatible 
with the advanced training scheme,
representation alignment.
\looseness=-1

\noindent\emph{RAE.}
We test the effectiveness of MixFlow
over the very recently developed strong model trained
with RAE~\cite{zheng2025rae}
that is different from most existing methods
and
replaces the VAE with pretrained representation encoders (\eg, DINO~\cite{oquabdinov2}) paired with trained decoders.
We post-train the officially-released model
using MixFlow
with $\gamma = 0.4$
for $200$ epochs.
The evaluation schemes are the same as RAE~\cite{zheng2025rae},
such as sampling steps ($50$),
auto-guidance and balanced sampling.
The auto-guidance scale is $1.5$.

The results are given in~\Cref{tab:GVPREPARAE}.
One can see that
the overall performance gets improved except 
that the IS scores are a little worse.
The gFID scores
are improved 
from $1.51$ to $1.43$
for generation without guidance
and from $1.13$ to $1.10$
for generation with guidance~\cite{Karras2024autoguidance}.
The superior performance
shows that
MixFlow is applicable
to the scenario:
learning diffusion models with the representation autoencoder paired with
trained decoder,
in addition to the widely-used VAE
encoder and decoder.

\noindent\textbf{Comparison to methods
to exposure bias alleviation.}
We compare MixFlow against
the modified training schemes, 
input perturbation~\cite{ning2023input},
and the modified inference schemes,
Epsilon Scaling~\cite{ning2023epsilonscaling},
and Time-Shift sampler~\cite{li2023timeshift}.
We implement these algorithms
and do the post-training process
with carefully hyperparameter tuning (details are provided in Appendix~\ref{appendix:moreresults})
in the SiT code base.

\begin{table*}[!th]
\footnotesize
\setlength{\tabcolsep}{1pt} 
\caption{\textbf{Comparison 
to representative methods
to exposure bias alleviation.}
The results are based on SiT-B.
} 
\vspace{-2mm}
\label{tab:comparewithexistingalgorithms}
\begin{tabular*}{\textwidth}{@{} l @{\extracolsep{\fill}}   c c c c c  c c c c c@{}}
\toprule
 \multirow{2}{*}{\textbf{ Method }}&   \multicolumn{5}{c}{\textbf{Generation  w/o guidance}} & \multicolumn{5}{c}{\textbf{Generation  w/ guidance}} \\
 \cmidrule(lr){2-6} \cmidrule(lr){7-11}
& gFID $\downarrow$ & sFID$\downarrow$ & IS $\uparrow$ & Pre. $\uparrow$ & Rec. $\uparrow$ & gFID $\downarrow$ & sFID $\downarrow$ & IS $\uparrow$ & Pre. $\uparrow$ & Rec. $\uparrow$ \\
\midrule
SiT-B/2~\cite{ma2024sit} & 17.97&6.43& 79.4 &0.62&\textbf{0.66} &4.46 &4.93 & 182.4 &	0.78&	\textbf{0.57}\\
\midrule
+ Time-Shift~\cite{li2023timeshift} & 18.12 & 6.41 & 79.5 & 0.62 & \textbf{0.66} & 4.49 & 4.80 & 180.5 & 0.78& 0.56\\
+ Epsilon Scaling~\cite{ning2023epsilonscaling} & 17.18 & 6.28 & 80.8 & 0.62 & 0.65 & 4.39 & 4.81 & 182.5 & 0.78 & 0.56\\
\arrayrulecolor{black!30}\midrule
+ Input Perturbation~\cite{ning2023input} & 17.01 & 5.37 & 83.1 & 0.62 & 0.65 & 4.32 & 4.77 & 185.8 & 0.78 & 0.56\\
\arrayrulecolor{black}\midrule
+ MixFlow  & \textbf{15.64} & \textbf{5.21} & \textbf{92.0} & \textbf{0.63} & \textbf{0.66} & \textbf{3.91} & \textbf{4.76} & \textbf{201.7} & \textbf{0.79} & 0.56  \\
\bottomrule
\end{tabular*}
\vspace{-4mm}
\end{table*}

\Cref{tab:comparewithexistingalgorithms} presents the comparison results.
One can see that
our approach MixFlow achieves
the most significant improvement.
The input perturbation algorithm
may include the ground-truth noisy data
at lower-noise timesteps
for training.
Such noisy data harms model training, which is validated in
\Cref{sec:ablation}
and \Cref{tab:sampledistributions} ($m_t \sim \mathcal{U}[0, 1]$).
Thus the perturbation strength needs
to be well tuned.
Otherwise, the performance gets worse.

The Time Shift sampler~\cite{li2023timeshift},
a modified inference scheme,
makes the timestep adjustment
for the next step sampling. 
The timestep adjustment may not be accurate
as the adjustment is heuristic,
and 
the adjusted timestep may correspond 
to the higher-noise timestep,
which implies that
it often needs 
more sampling steps
to reach the timestep $1$.
These are the reasons why this scheme
gets worse results.

Epsilon Scaling sampling~\cite{ning2023epsilonscaling}
is a method that moves the sampling trajectory 
closer to the vector field learned in the training phase by scaling the network output epsilon.
One can see that it can improve the performance
and it is inferior to our approach MixFlow.
Similar to
Time Shift, 
Epsilon Scaling is heuristic
and the adjustment may not be accurate. 
The results indicate that 
our training approach,
MixFlow,
is advantageous.

\begin{table}[t]
\centering
    \footnotesize    
    \setlength{\tabcolsep}{0.9pt} 
\caption{\textbf{The performance for text-to-image generation.}
}
\vspace{-2mm}
\label{tab:t2i}
\begin{tabular*}{\columnwidth}{@{} l @{\extracolsep{\fill}} cccccc @{}} 
\toprule
\multirow{2}{*}{\textbf{ Method }} & \multicolumn{2}{c}{GenEval} & \multicolumn{1}{c}{DPG-Bench} & \multicolumn{3}{c}{T2I-CompBench} \\
 \cmidrule(lr){2-3}  
 \cmidrule(lr){5-7}
    & Avg. $\uparrow$ & Counting $\uparrow$ & Avg. $\uparrow$ & Shape  $\uparrow$ & Spatial $\uparrow$  & Complex  $\uparrow$\\
    \midrule
    \multicolumn{7}{l}{\emph{$40$ sampling steps}}\\ 
\arrayrulecolor{black!30}\midrule
  SD $3.5$     & 0.64 & 0.50 & 84.80   & 0.6555 & 0.2850 & 0.4202\\
    SD $3.5$-ft-20k   & 0.64 & 0.51 & 84.90  & 0.6552 & 0.2852 & 0.4208\\
    SD $3.5$-ft-10k   & 0.64 & 0.51 & 84.87 & 0.6556 & 0.2855 & 0.4205\\
 + MixFlow     & \textbf{0.69} & \textbf{0.63} & \textbf{87.49}  & \textbf{0.7013} & \textbf{0.3642} & \textbf{0.4756}\\
  \arrayrulecolor{black} \midrule
\multicolumn{7}{l}{\emph{$10$ sampling steps}}\\ 
  \arrayrulecolor{black!30}\midrule
SD $3.5$     & 0.62 & 0.47 & 81.78 & 0.6344 & 0.2510& 0.3872\\
    SD $3.5$-ft-20k   & 0.62 & 0.48 & 81.83 &0.6348 & 0.2532 & 0.3878 \\

SD $3.5$-ft-10k   & 0.62 & 0.48 & 81.80 & 0.6349 & 0.2530 & 0.3874\\
 + MixFlow    & \textbf{0.67} & \textbf{0.60} &  \textbf{85.88}  & \textbf{0.6987} & \textbf{0.3318} &\textbf{0.4223}\\
  \arrayrulecolor{black}\bottomrule
\end{tabular*}
\vspace{-7mm}
\end{table}

\noindent\textbf{Improvement over the text-to-image generation model SD $3.5$.}
We demonstrate MixFlow
on the text-to-image (T2I) generation task
over the Stable Diffusion
$3.5$ model (SD $3.5$-medium)~\cite{esser2024scaling}.
We build a high-quality T2I dataset,
and finetune the model 
with the SD $3.5$ training scheme
for $10K$ steps,
which is enough for convergence.
We name the resulting model as
SD $3.5$-ft-$10$K.
We post-train this model
with the MixFlow training algorithm 
for $10K$ steps
using the  
training hyperparameters
same as SD $3.5$.
We build a baseline model,
SD $3.5$-ft-$20$K,
by
further post-training 
SD $3.5$-ft-$10$K
for extra $10$K steps,
to show that our superior results
do not come from the extra training steps.
\Cref{tab:t2i}
gives the results
over three metrics.
The results for example sub-metrics are also included.
Detailed results on more sub-metrics are in
Appendix~\ref{appendix:moreresults}
and Table~\ref{tab:t2icompbench}.
One can see that
the MixFlow training
gets improved
in terms of
all the six metrics
for both $40$ and $10$
sampling steps.
The improvement
shows that
MixFlow 
is applicable
to the text-to-image generation model,
where the
complex network structure MMDiT
is used
and the pretrained model is
trained using advanced schemes.

\begin{figure}[t]
    \centering
\footnotesize
    \begin{subfigure}{\linewidth}
        \centering
        \includegraphics[width=0.32\linewidth]{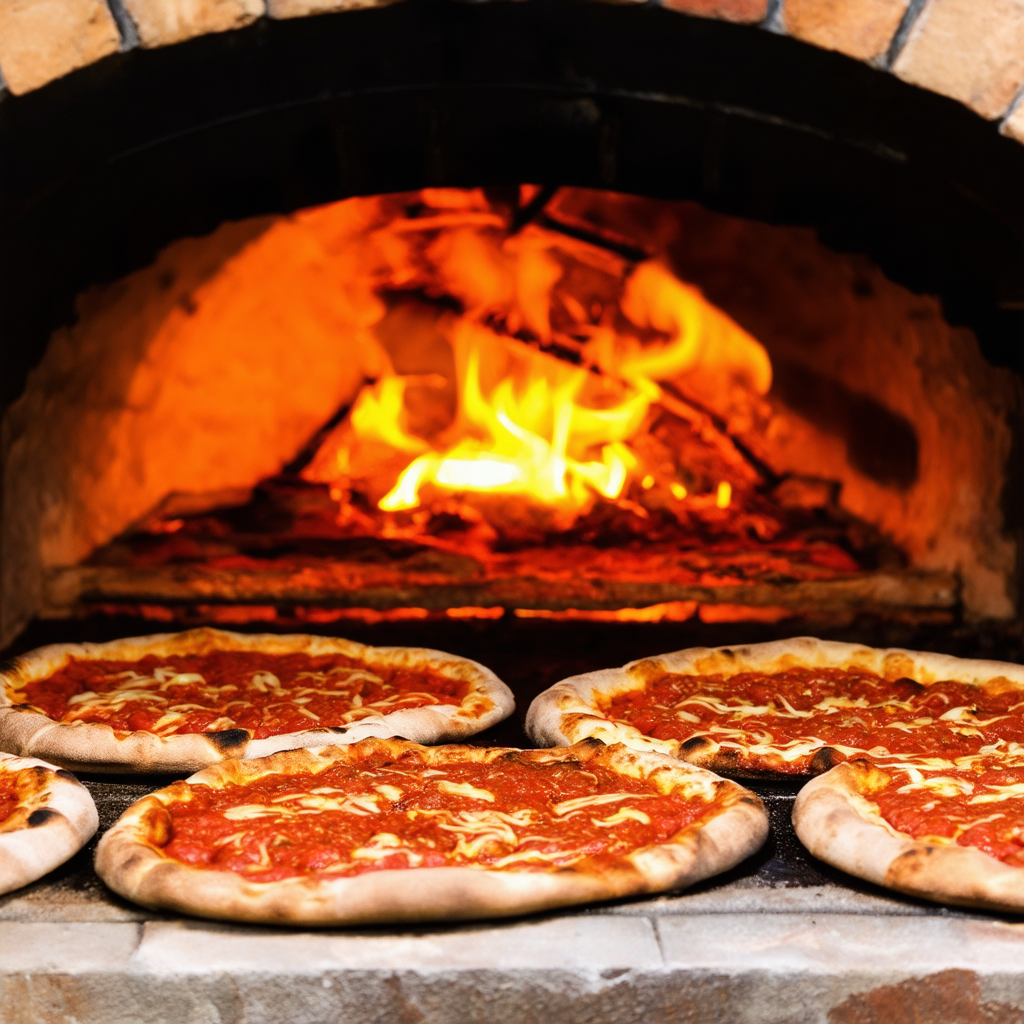} \hfill
        \includegraphics[width=0.32\linewidth]{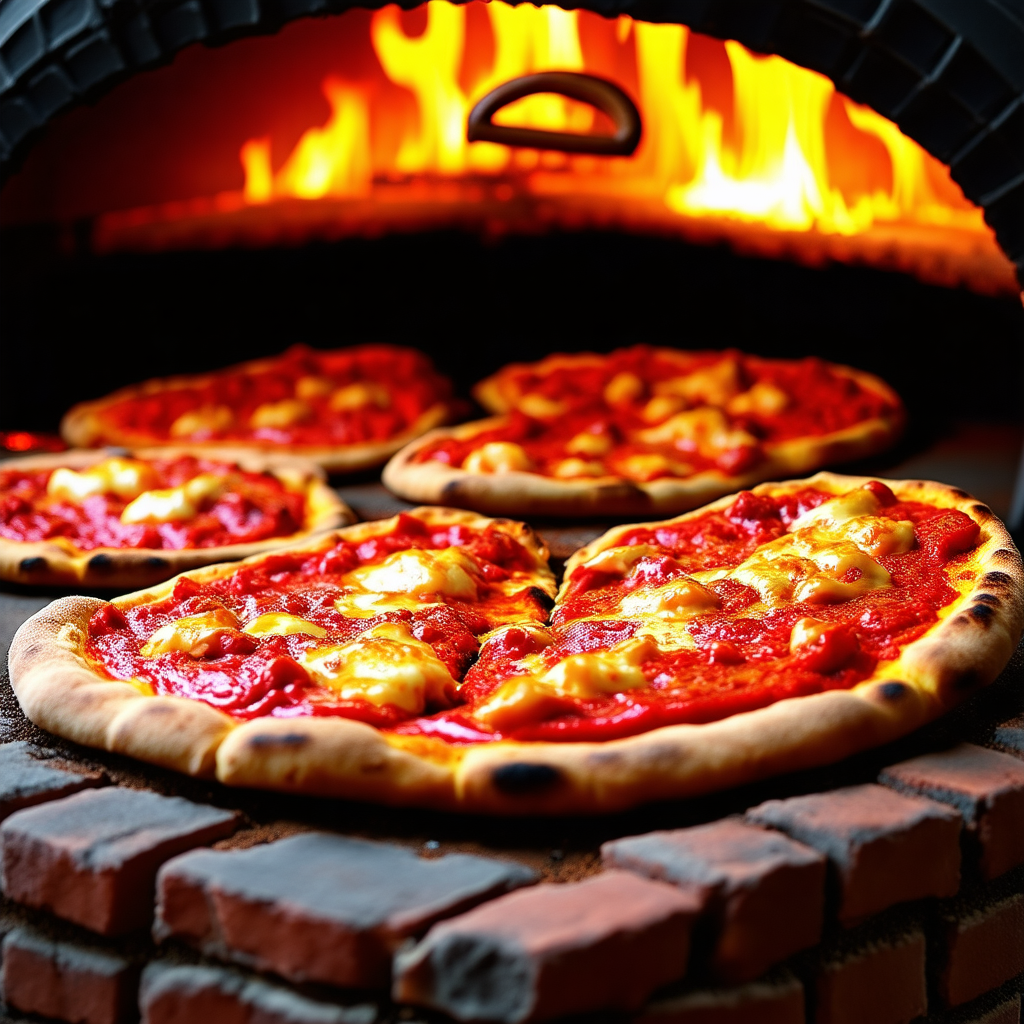} \hfill
        \includegraphics[width=0.32\linewidth]{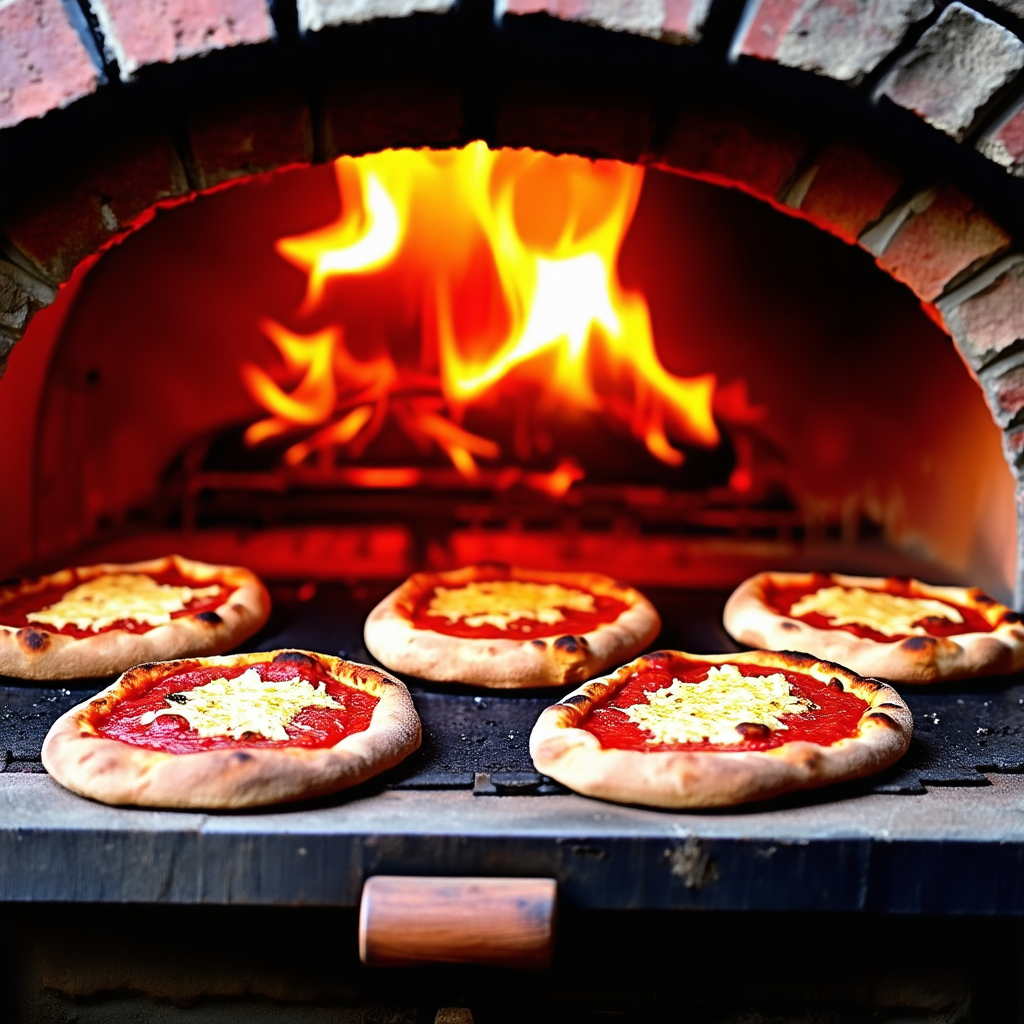}
        \caption{\textbf{Five} savory pizzas with vibrant red tomato sauce and golden cheese...}
        \label{fig:row_pizzas}
    \end{subfigure}

    \begin{subfigure}{\linewidth}
        \centering
        \includegraphics[width=0.32\linewidth]{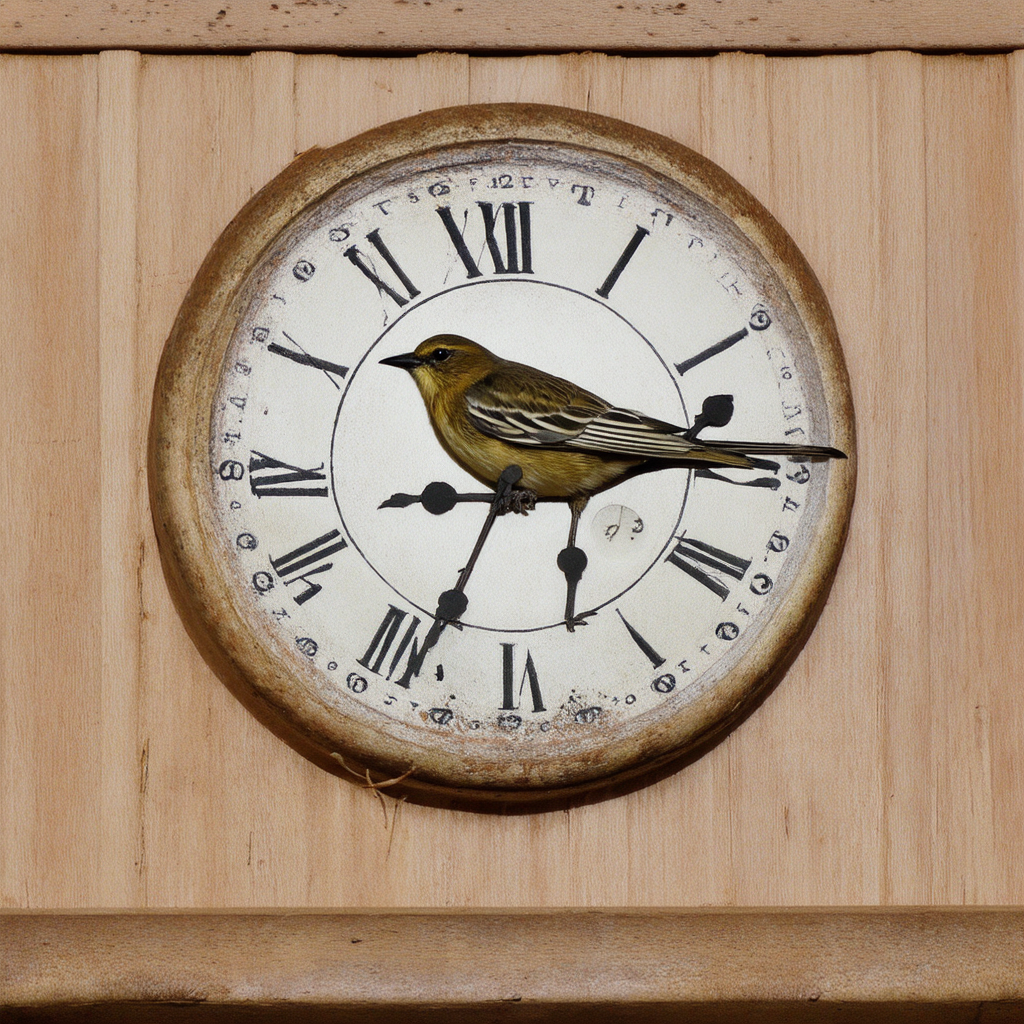} \hfill
        \includegraphics[width=0.32\linewidth]{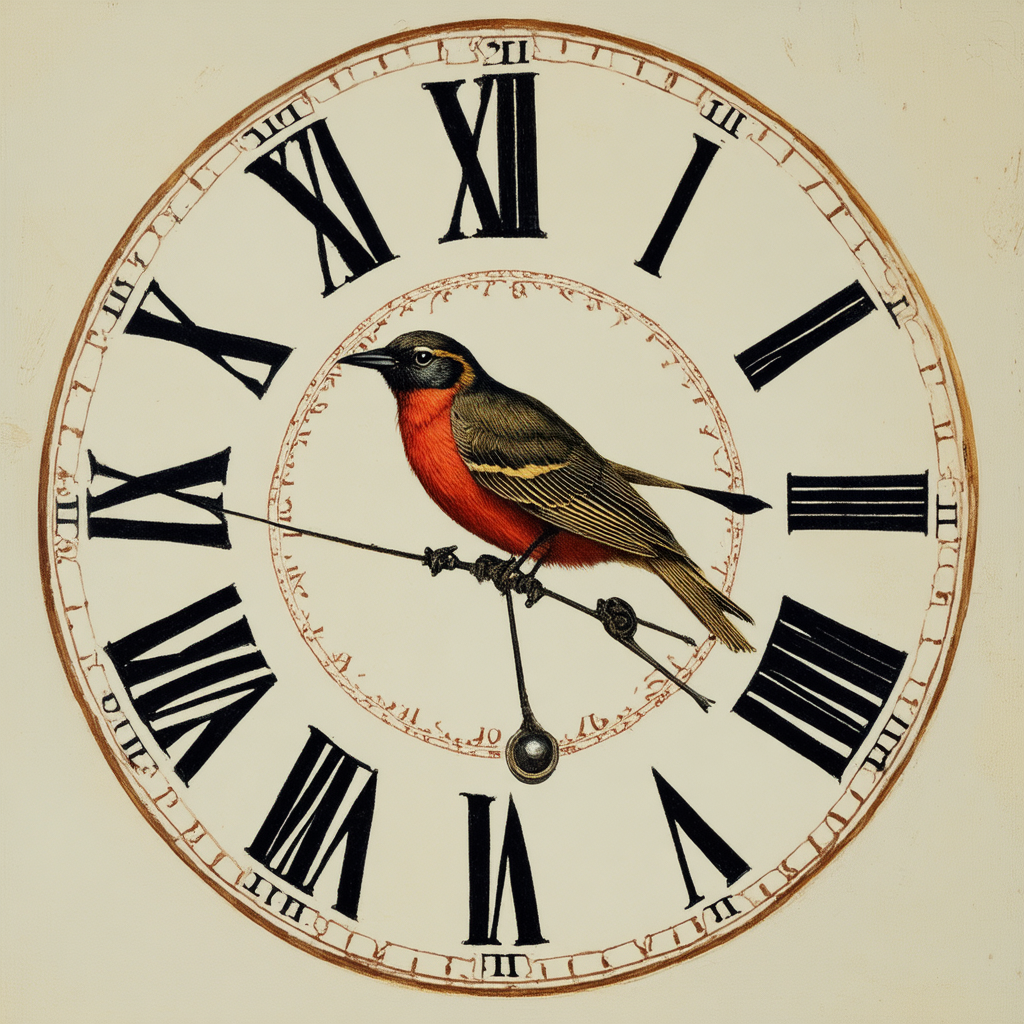} \hfill
        \includegraphics[width=0.32\linewidth]{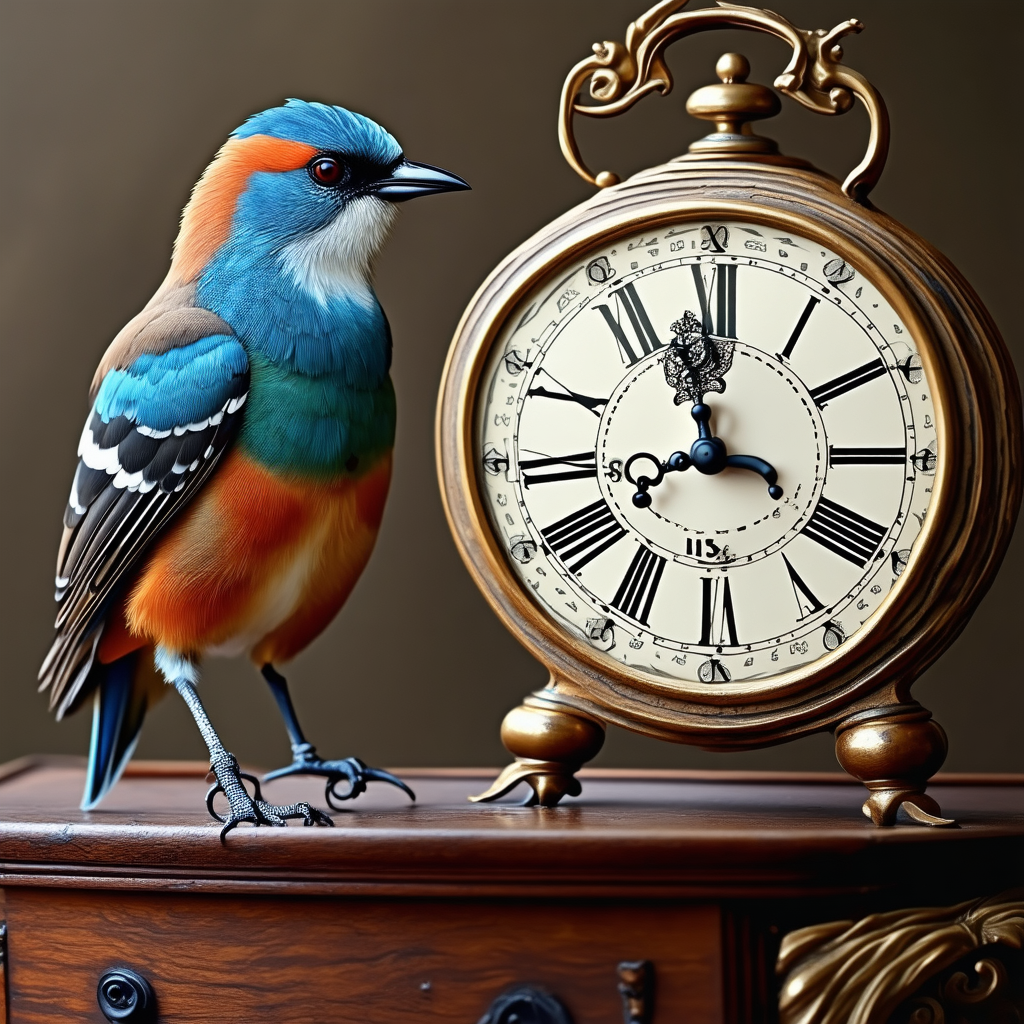}
        \caption{A bird on the \textbf{left} of a clock...}
        \label{fig:row_asparagus}
    \end{subfigure}

    \begin{subfigure}{\linewidth}
        \centering
        \includegraphics[width=0.32\linewidth]{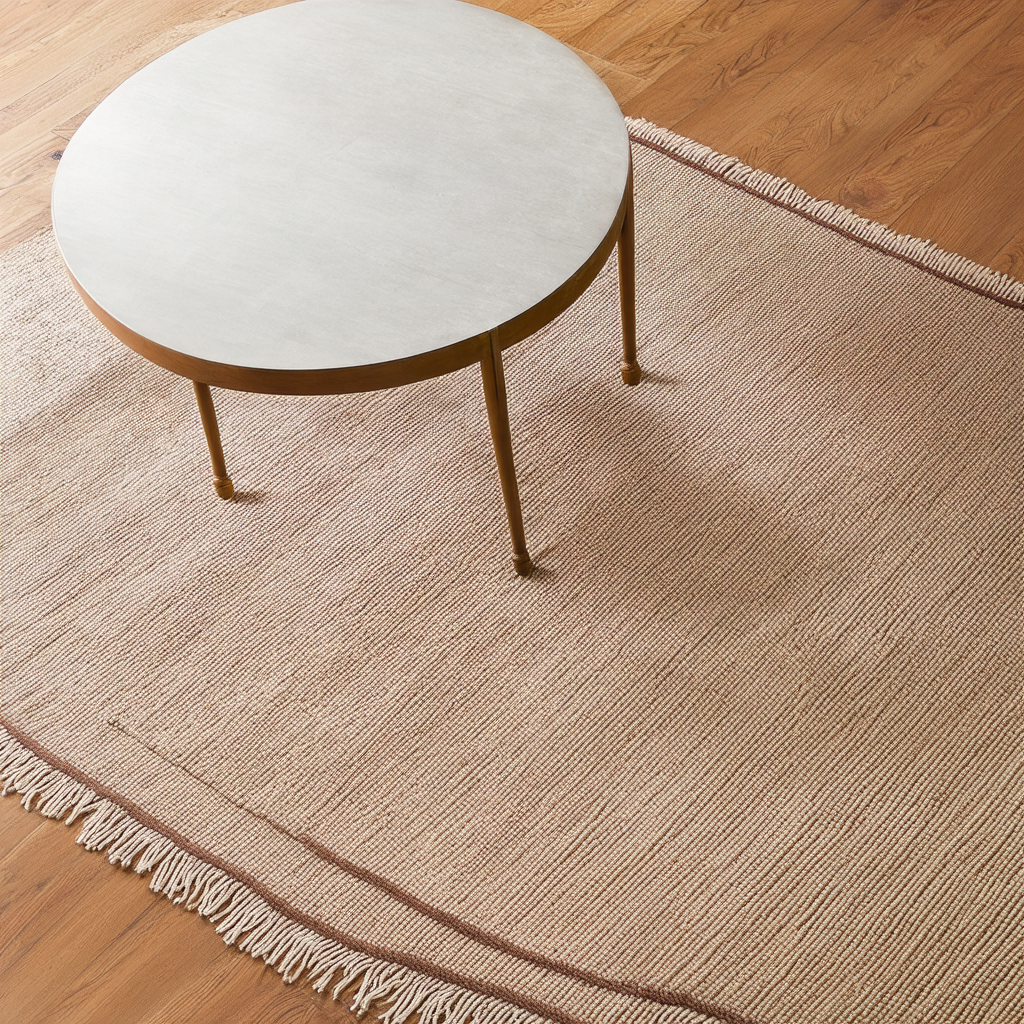} \hfill
        \includegraphics[width=0.32\linewidth]{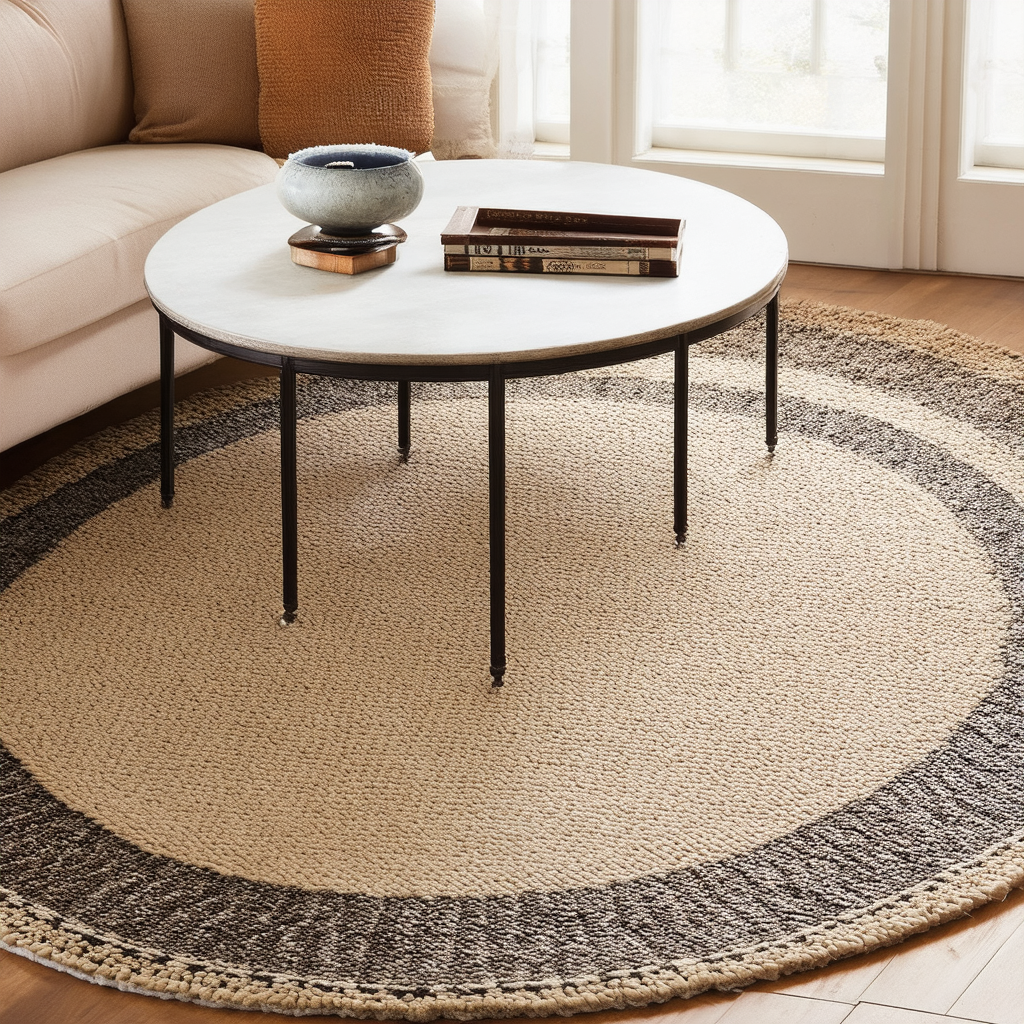} \hfill
        \includegraphics[width=0.32\linewidth]{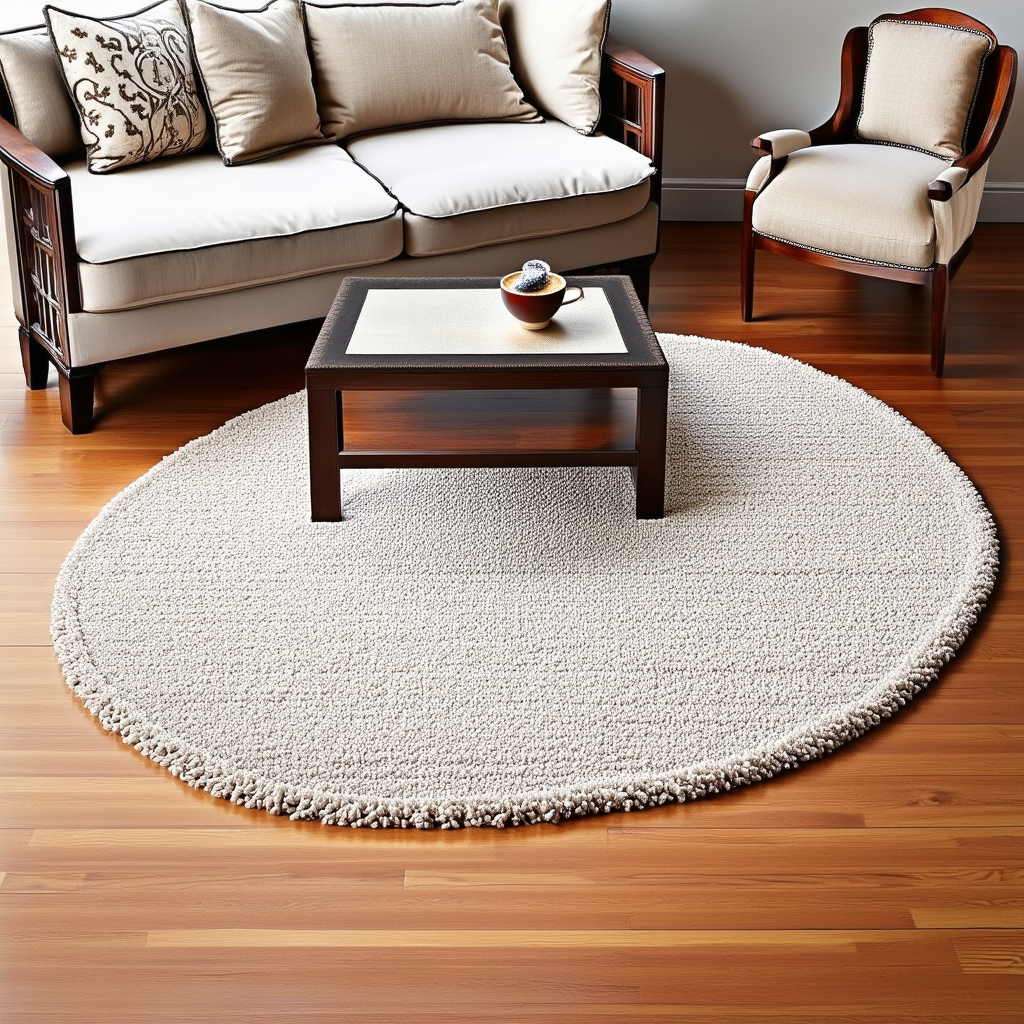}
        \caption{An oval rug and a \textbf{square} coffee table...}
        \label{fig:shape}
    \end{subfigure}
    \vspace{-6mm}
    \caption{ Example results illustrating
    the advantages of MixFlow
    with respect to counting, spatial relationship and object shape.
    Left: SD 3.5;
    Middle: SD 3.5-ft-20k;
    Right: MixFlow.}
    \label{fig:visual_comparison}
\vspace{-7mm}
\end{figure}

\Cref{fig:visual_comparison}
shows visual results
that demonstrate the advantage 
of MixFlow
for three 
example sub-metrics:
counting, spatial relation,
and shape.
The results are from $40$
sampling steps.
For example,
from
\Cref{fig:visual_comparison} (b),
one can see that our approach
can follow the prompt
to generate the image
with a bird really on the left of a clock.

\begin{table*}[t]
\centering
\label{tab:imagenet256}
\footnotesize
\setlength{\tabcolsep}{7pt}
\caption{\textbf{
Class-conditional performance on ImageNet $256 \times 256$}. 
Our approach MixFlow achieves the best gFID score without guidance (1.43) and with guidance (1.10).
We follow RAE~\cite{zheng2025rae} to use class-balanced sampling for some results, which 
are indicated in gray. 
The results for SiT, REPA, REPA + MixFlow and SiT+MixFlow
are from the SDE sampler.
The top $3$ results per metric are marked in \textbf{bold}.}
\vspace{-2mm}
\label{tab:sotaresults256}
\begin{tabular}{@{}lrr rrrrr rrrrr@{}}
\toprule
\multirow{2}{*}{\textbf{ Method }} & \multicolumn{1}{c}{\textbf{Epochs}} & \multicolumn{1}{c}{\textbf{\#Params}} & \multicolumn{5}{c}{\textbf{Generation  w/o guidance}} & \multicolumn{5}{c}{\textbf{Generation  w/ guidance}} \\
\cmidrule(lr){4-8} \cmidrule(lr){9-13}
 & & & \multicolumn{1}{c}{gFID$\downarrow$} &
 \multicolumn{1}{c}{sFID$\downarrow$} &
 \multicolumn{1}{c}{IS$\uparrow$} & \multicolumn{1}{c}{Prec.$\uparrow$} & \multicolumn{1}{c}{Rec.$\uparrow$} & \multicolumn{1}{c}{gFID$\downarrow$} & 
 \multicolumn{1}{c}{sFID$\downarrow$} &
 \multicolumn{1}{c}{IS$\uparrow$} & \multicolumn{1}{c}{Prec.$\uparrow$} & \multicolumn{1}{c}{Rec.$\uparrow$} \\
\midrule
\multicolumn{11}{@{}l}{\textit{Autoregressive}} \\
\arrayrulecolor{black!30}\midrule
\rowcolor{black!5}VAR~\cite{VAR}& 350 & 2.0B & – & – & – & – & –  &  1.73 & – & \textbf{350.2} & 0.82 & 0.60 \\
\rowcolor{black!5}MAR~\cite{MAR} & 800 & 943M & 2.35 & – & 227.8 & \textbf{0.79} & 0.62 & 1.55  & & 303.7 & 0.81 & 0.62 \\
\rowcolor{black!5}xAR~\cite{XAR} & 800 & 1.1B & – & – & – & – & –& 1.24  & – & 301.6 & \textbf{0.83} & 0.64 \\
\arrayrulecolor{black}\midrule
\multicolumn{11}{@{}l}{\textit{Diffusion}} \\
\arrayrulecolor{black!30}\midrule
ADM~\cite{ADM}  & 400 & 554M & 10.94 & 6.02 &   101.0 & 0.69 & 0.63 & 3.94 & 6.14& 215.8 & \textbf{0.83} & 0.53 \\
RIN~\cite{RIN} & 480 & 410M & 3.42& – & 182.0 & – & – & – & – & – & – & – \\
PixNerd~\cite{wang2025pixnerd} & 160 & 700M & – & – & – & – & –& 1.87 & 4.36 & 298.0 &  0.79 &  0.61 \\
SiD2~\cite{hoogeboom2024simpler} & 1280 & – & – &– & – & – & – & 1.38 & – &– & – & – \\
DiT~\cite{peebles2023DiT} & 1400 & 675M & 9.62 & 6.85 & 121.5 & 0.67 & 0.67 & 2.27 & 4.60 & 278.2 & \textbf{0.83} & 0.57 \\
MaskDiT~\cite{zheng2023fast} & 1600 & 675M & 5.69 & 10.34 & 177.9 & 0.74 & 0.60 & 2.28 & 5.67 & 276.6 & 0.80 & 0.61 \\
SiT~\cite{ma2024sit} & 1400 & 675M & 8.61 & 6.32 & 131.7 & 0.68 & 0.67 & 2.06 & 4.50 & 270.3 & 0.82 & 0.59 \\
MDTv2~\cite{gao2023mdtv2} & 1080 & 675M & – & – & – & – & – & 1.58 & 4.52 & \textbf{314.7} & 0.79 & 0.65 \\
REG~\cite{wu2025representation} & 800 & 677M&  1.80 &\textbf{4.59}& \textbf{230.8}& 0.77& 0.66 &1.36 & \textbf{4.25}&  299.4 & 0.77 & 0.66\\
\rowcolor{black!5}REPA ~\cite{yu2025repa} & 800 & 675M & 5.84 & 5.79 & 158.7 & 0.70 & \textbf{0.68} & 1.28 & 4.68 & 305.7 & 0.79 & 0.64 \\
\rowcolor{black!5}VA-VAE ~\cite{yao2025reconstruction}& 800 &675M & 2.05 & \textbf{4.37} &  207.7 & 0.77 & 0.66 & 1.25 & \textbf{4.15} & 295.3 & 0.80 & \textbf{0.65} \\
\rowcolor{black!5}REPA-E~\cite{Leng_2025_ICCV} & 800 & 675M & \textbf{1.69} & \textbf{4.17}& 219.3 & 0.77 & 0.66 & \textbf{1.12} & \textbf{4.09} & 302.9 & 0.79 & 0.66 \\
\rowcolor{black!5}DDT ~\cite{wang2025ddt} & 400 & 675M & 6.27 &  – & 154.7 & 0.68 & \textbf{0.69} & 1.26 &  – & 310.6 & 0.79 & 0.65 \\
\rowcolor{black!5}RAE~\cite{zheng2025rae} &  800 & 839M  & \textbf{1.51} & 5.31 & \textbf{242.9} & \textbf{0.79} & 0.63 & \textbf{1.13} & 4.74 & 262.6 & 0.78 & \textbf{0.67} \\
\arrayrulecolor{black}\midrule
MixFlow + SiT-XL & 200 & 675M &  7.56 & 4.75 & 144.5 & 0.69 & \textbf{0.68} & 1.97 & 4.34 & 276.8 & 0.82 & 0.60 \\
\rowcolor{black!5}MixFlow + REPA & 200 & 675M & 5.00 & 4.87 & 171.4 & 0.72 & 0.67 & 1.22 & 4.49 & \textbf{313.4} & 0.80 & 0.64    \\
\rowcolor{black!5}MixFlow + RAE & 200 & 839M & \textbf{1.43} & 4.81 & \textbf{239.8} & \textbf{0.80} & 0.64 & \textbf{1.10} & 4.40 & 259.7 & 	0.78	& \textbf{0.67} \\
\arrayrulecolor{black}\bottomrule
\end{tabular}%
\vspace{-2mm}
\end{table*}

\begin{table}[]
\centering
\footnotesize
\setlength{\tabcolsep}{7pt}
\caption{\textbf{Class-conditional performance on ImageNet $512 \times 512$} with guidance.
}
\vspace{-2mm}
\label{tab:sotaresults512}
\begin{tabular}{lrrrrr}
\toprule
\textbf{ Method } 
 & gFID$\downarrow$ & sFID$\downarrow$ & IS$\uparrow$ & Prec.$\uparrow$ & Rec.$\uparrow$ \\
\midrule
\multicolumn{6}{@{}l}{\textit{Autogressive}} \\
\arrayrulecolor{black!30}\midrule

MAGViT-v2~\cite{yu2023language}& 1.91  &–& \textbf{324.3} & –  & –  \\
\rowcolor{black!5} VAR~\cite{VAR} & 2.63 & – & 303.2 & –  & –  \\
\rowcolor{black!5} XAR~\cite{XAR} & 1.70 & –& 281.5 & –  & –  \\
\arrayrulecolor{black}\midrule
\multicolumn{6}{@{}l}{\textit{Diffusion}} \\
\arrayrulecolor{black!30}\midrule
ADM~\cite{ADM} & 3.85 & 5.86 & 221.7 & \textbf{0.84} & 0.53 \\
SiD2~\cite{hoogeboom2024simpler} & 1.50 & –  & –  & –  & –  \\
DiT~\cite{peebles2023DiT} & 3.04 &5.02 &240.8 & \textbf{0.84} & 0.54 \\
EDM2~\cite{karras2024analyzing} & \textbf{1.25} & –  & –  & –  & –  \\
SiT~\cite{ma2024sit} & 2.62 & \textbf{4.18} &252.2 & \textbf{0.84} & 0.57 \\
DiffiT~\cite{hatamizadeh2024diffit} & 2.67 & –  &252.1 & 0.83 & 0.55 \\
REPA~\cite{yu2025repa}& 2.08 & 4.19 &274.6 & 0.83 & 0.58 \\
REG~\cite{wu2025representation} & 1.68& \textbf{3.87} &  \textbf{306.9} & 0.80 & \textbf{0.63} \\
\rowcolor{black!5} 
DDT~\cite{wang2025ddt} & 1.28 & 4.22 & \textbf{305.1} & 0.80 & \textbf{0.63} \\
\rowcolor{black!5} 
RAE~\cite{zheng2025rae} & \textbf{1.13} & 4.24 & 259.6 & 0.80 & \textbf{0.63} \\
\arrayrulecolor{black}\midrule
\rowcolor{black!5} MixFlow + RAE & \textbf{1.10} & \textbf{4.17} & 256.9 & 0.80 & \textbf{0.65}\\
 
\arrayrulecolor{black}\bottomrule
\end{tabular}
\vspace{-4mm}
\end{table}

\subsection{State-of-the-Art
Class-Conditional Generation}
\label{sec:sotaresults}
We present the comparison
with state-of-the-art models on ImageNet 
$256 \times 256$ and $512 \times 512$.
We follow RAE~\cite{zheng2025rae}
to use auto-guidance and class-balanced sampling for evaluation.
The guidance scale is $1.5$ for both $256 \times 256$ and $512 \times 512$,
which is the same as RAE for $512 \times 512$.

\Cref{tab:sotaresults256,tab:sotaresults512}
present the results.
The results without guidance for ImageNet $512 \times 512$ are included in Appendix~\ref{appendix:sotawoguidance512}.
MixFlow + RAE gets superior performance over all prior diffusion models  in terms of gFID, setting new state-of-the-art gFID scores of $1.43$ without guidance 
and $1.10$ with guidance at $256 \times 256$,
$1.55$ without guidance
and $1.10$ with guidance at $512 \times 512$.

Our state-of-the-art FID score is supported by strong qualitative results, which demonstrate high semantic diversity and fine-grained details. Visualization samples are provided 
in Appendix~\ref{appendix:visualresultsimagenet}.
\vspace{-1mm}
\section{Conclusion}
\label{sec:conclusion}
\vspace{-1mm}
We present a MixFlow training approach
to alleviating exposure bias
for diffusion models.
The key is to exploit 
slowed interpolations,
\ie, higher-noise interpolations,
for training the prediction network
at each training timestep.
We demonstrate the effectiveness
on various generation models.
The MixFlow + RAE models achieve strong generation results on ImageNet: 
$1.43$ FID (without guidance)
and $1.10$ (with guidance) at $256 \times 256$,
and $1.55$ FID (without guidance)
and $1.10$ (with guidance) at $512 \times 512$.

\noindent\textbf{Acknowledgments.} This work was supported by the Shanghai Oriental Talents Program (Youth Project) (No. QNKJ2024060), the Natural Science Foundation of Shanghai (No. 24ZR1407200), and the Shanghai Municipal Commission of Economy and Informatization (No. 2025-GZL-RGZN-BTBX-01011).

\small
\bibliographystyle{ieeenat_fullname}
\bibliography{main}

\clearpage

\counterwithin{figure}{section}
\counterwithin{table}{section}
\appendix


\section{Details for Figure~\ref{fig:slowflow}}
\label{appendix:detailsforfigure1}

\begin{figure}[b]
  \centering
  \footnotesize
\subfloat[Standard Training]{\includegraphics[width=0.47\linewidth]{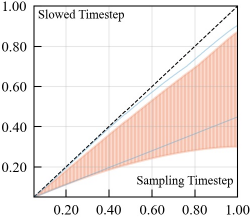}}
~~~~
\subfloat[MixFlow Training]{\includegraphics[width=0.47\linewidth]{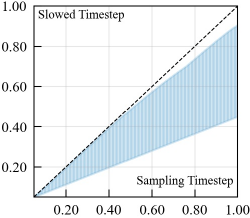}}
  \caption{
   \textbf{Slow Flow for diffusion models.}
   The upper and lower envelopes of the shading area in (b) are shown in (a).
}
 \label{fig:slowflow_gvp}
\end{figure}

\noindent\textbf{Implementation.}
The ground truth noisy data
that is the nearest to
the generated noisy data $\hat{\mathbf{x}}$
is the projected data
of the generated noisy
data
along the velocity
$\mathbf{x}_1 - \mathbf{x}_0$ for flow matching.
The corresponding slowed timestep
is computed
as:
\begin{align}
    m_t = \frac{(\hat{\mathbf{x}}_t - \mathbf{x}_0)^\top (\mathbf{x}_1 - \mathbf{x}_0)}{\|\mathbf{x}_1 - \mathbf{x}_0\|_2^2}.
    \label{eq:projection}
\end{align}

We perform the ODE sampling on $20{,}000$ training images 
on the SiT-B model 
for $50$ sampling steps from $t=0.05$ to $t=1.0$.
For each sampling step $t$, we collect the slowed timesteps $m_t$ across all $20{,}000$ images.
The shaded area 
represents the range of slowed timesteps at each sampling timestep.
At each sampling timestep,
the lower point is the minimum slowed timestep, and the upper point is the maximum slowed timestep among all the $20{,}000$ images.

\vspace{1mm}
\noindent\textbf{Slow Flow for diffusion with GVP.}
\Cref{fig:slowflow_gvp} shows 
the Slow Flow phenomenon 
for diffusion with generalized variance preserving interpolations the on SiT-B model~\cite{ma2024sit}.
Different from flow matching
that uses linear interpolation,
we find the nearest ground truth noisy data
by checking $1000$ noisy data samples
for each generated noisy data
and record the corresponding slowed timestep.
The observation
is consistent to that
for flow matching
in Figure~\ref{fig:slowflow}
in the main paper.

\section{Details for Figure~\ref{fig:toyexample}}
\label{appendix:detailsforfigure3}
The distribution for noise $x_0$ is a Gaussian: $p(x_0) = \mathcal{N}(x_0; 0, 1)$.
The distribution for data $x_1$ is a mixture of two Gaussians: $p(x_1) = 0.5 \mathcal{N}(x_1; -2, 0.1^2) + 0.5 \mathcal{N}(x_1; 2, 0.1^2)$.

The velocity prediction network is an MLP with $4$ hidden layers,
$256$ hidden dimensions,
and the Swish activation function.
The optimizer is Adam with learning rate $0.001$.
The batch size is $2{,}048$.
The training iteration number
is $26{,}000$ for standard training,
and $6{,}000$ from the model trained with $20{,}000$ iterations
for MixFlow training.

We use the Euler method with $5$ sampling steps.
We transform the generated noisy data to a distribution, by using Kernel Density Estimation (KDE) as implemented in the SciPy library. We utilize the default hyperparameter configuration: employ a Gaussian kernel with the bandwidth determined via Scott’s Rule.

\section{More Results
for Ablation Studies (Section~\ref{sec:ablation})}
\label{appendix:moreresults}

\begin{table*}[t]
\footnotesize
\setlength{\tabcolsep}{1pt} 
\caption{\textbf{The performance with
the SDE sampler}
for class-conditional generation on ImageNet 
$256 \times 256$ and $512 \times 512$.}
\begin{tabular*}{\textwidth}{@{} l  @{\extracolsep{\fill}}  c c c c c  c c c c c@{}}
\toprule
  \multirow{2}{*}{\textbf{ Method }} &  \multicolumn{5}{c}{\textbf{Generation  w/o guidance}} & \multicolumn{5}{c}{\textbf{Generation  w/ guidance}} \\
 \cmidrule(lr){2-6} \cmidrule(lr){7-11}
    & gFID $\downarrow$ & sFID$\downarrow$ & IS $\uparrow$ & Pre. $\uparrow$ & Rec. $\uparrow$ & gFID $\downarrow$ & sFID $\downarrow$ & IS $\uparrow$ & Pre. $\uparrow$ & Rec. $\uparrow$ \\
\midrule
\multicolumn{11}{l}{\emph{ImageNet $256 \times 256$}}\\
\arrayrulecolor{black!30}\midrule
SiT-B/2  & 17.19 & 6.51 & 82.9 & 0.63 & 0.66 & 4.10 & 4.98 & 192.4 & 0.79 & 0.56 \\
+ MixFlow  & 14.47 & 5.23 & 96.3 &0.65 & 0.65 & 3.74 & 4.52 & 212.5 & 0.80 & 0.54  \\
\arrayrulecolor{black!30}\midrule
SiT-XL/2   & 8.26 & 6.32  & 131.7 & 0.68 & 0.67 & 2.06 & 4.60 & 270.3 & 0.83 & 0.59\\
+ MixFlow   &  7.56 & 4.75 & 144.5 & 0.69 & 0.67 & 1.97 & 4.34 & 276.8 & 0.82 & 0.60  \\
\arrayrulecolor{black}\midrule
\multicolumn{11}{l}{\emph{ImageNet $512 \times 512$}}
\\
\arrayrulecolor{black!30}\midrule
SiT-XL/2  & 9.20  & 7.33  & 125.3 & 0.78 & 0.64 & 2.62 & 4.18 & 252.2 & 0.84 & 0.57\\
+ MixFlow & 8.46& 5.76 & 139.7 & 0.79  & 0.64 & 2.49 & 4.01 & 268.9 & 0.84 & 0.57\\
\arrayrulecolor{black}\bottomrule
\end{tabular*}
\label{tab:sitflowmatchingSDEsolver}
\end{table*}

We present
more results in terms of other metrics for ablation studies.

\vspace{1mm}
\noindent\textbf{Sampling distributions for $t$ and $m_t$.}
\Cref{tab:moreresultsonsamplingdistributions} presents the results for other
metrics,
including
sFID, IS, Precision, and Recall.
The overall observations are consistent
to that for the gFID metric.

\begin{table*}[!th]
\footnotesize
\setlength{\tabcolsep}{1pt} 
\caption{\textbf{Studies of
    sampling distributions 
    for $t$ and $m_t$ in 
    the MixFlow training.} } 
\label{tab:moreresultsonsamplingdistributions}
\begin{tabular*}{\textwidth}{@{} l  @{\extracolsep{\fill}}  c c c c c  c c c c c@{}}
\toprule
 &  \multicolumn{5}{c}{\textbf{Generation  w/o guidance}} & \multicolumn{5}{c}{\textbf{Generation  w/ guidance}} \\
 \cmidrule(lr){2-6} \cmidrule(lr){7-11}
    & gFID $\downarrow$ & sFID$\downarrow$ & IS $\uparrow$ & Pre. $\uparrow$ & Rec. $\uparrow$ & gFID $\downarrow$ & sFID $\downarrow$ & IS $\uparrow$ & Pre. $\uparrow$ & Rec. $\uparrow$ \\
\midrule

$m_t \sim \mathcal{U}[0,t], t\sim \mathcal{U}[0,1]$ & 16.57 & 5.30 & 88.4 & 0.62 & 0.64 & 4.25 & 4.77& 195.9 & 0.78 & 0.55 \\

$m_t \sim \mathcal{U}[0,1], t\sim \mathcal{U}[0,1]$ & 18.27 & 5.74 & 75.8 & 0.62 & 0.66 & 5.07 & 4.89& 168.9 & 0.77 & 0.56 \\
\arrayrulecolor{black!30}\midrule

$m_t \sim \mathcal{U}[0,t], t \sim \operatorname{Beta}(2,1)$ &  15.64 & 5.31 & 90.5 & 0.63 & 0.64 & 3.93 & 4.74 & 201.8 & 0.78 & 0.55    \\

$m_t \sim \mathcal{U}[0,1], t \sim \operatorname{Beta}(2,1)$ &    18.25 & 5.73 & 76.3 & 0.62 & 0.65 & 5.06 & 4.88 & 168.8 & 0.77 & 0.57\\

\arrayrulecolor{black}\bottomrule
\end{tabular*}
\end{table*}

\vspace{1mm}
\noindent\textbf{Mixture range coefficient $\gamma$ in the distribution
$\mathcal{U}[(1-\gamma)t, t]$
for sampling $m_t$.}
\Cref{table:delta_magnitude} presents the results for sFID, IS, Precision, and Recall for $\gamma \in \{0.0, 0.3, 0.5, 0.7, 0.8, 0.9, 1.0\}$.
The observations are also consistent
to that for the gFID metric.

\begin{table*}[!th]
\footnotesize
\setlength{\tabcolsep}{1pt} 
\caption{\textbf{Studies of sampling distributions of $m_t$.}
 $m_t \sim \mathcal{U}[(1-\gamma)t, t]$.
} 
\label{table:delta_magnitude}
\begin{tabular*}{\textwidth}{@{} l  @{\extracolsep{\fill}}  c c c c c  c c c c c@{}}
\toprule
 &  \multicolumn{5}{c}{\textbf{Generation  w/o guidance}} & \multicolumn{5}{c}{\textbf{Generation  w/ guidance}} \\
 \cmidrule(lr){2-6} \cmidrule(lr){7-11}
    & gFID $\downarrow$ & sFID$\downarrow$ & IS $\uparrow$ & Pre. $\uparrow$ & Rec. $\uparrow$ & gFID $\downarrow$ & sFID $\downarrow$ & IS $\uparrow$ & Pre. $\uparrow$ & Rec. $\uparrow$ \\
\midrule
SiT-B/2 & 17.97&6.43& 79.4 &0.62&0.66 &4.46 &4.93 & 182.4 &	0.78&	0.57\\
\midrule
$\gamma = 0.0$ & 18.00& 6.45 & 79.1 & 0.62 & 0.66 & 4.48 & 5.01 & 181.1 & 0.78 & 0.57\\
$\gamma = 0.3$ & 15.75 & 5.28 & 84.1 & 0.63 & 0.65 & 4.09 & 5.04 & 195.7 & 0.79 & 0.56\\
$\gamma = 0.5$ & 15.72 & 5.20 & 87.8 & 0.63 & 0.65 & 4.00 & 4.80 & 197.2 & 0.79 & 0.56\\
$\gamma = 0.7$ & 15.65 & 5.22 & 90.2 & 0.63 & 0.66 & 3.95 & 4.76 & 200.6& 0.79 & 0.56 \\
$\gamma = 0.8$ & 15.64 & 5.21 & 92.0 & 0.63 & 0.66 & 3.91 & 4.76 & 201.7 & 0.79 & 0.56   \\
$\gamma = 0.9$ & 15.64& 5.24 & 91.0 & 0.63 & 0.66 & 3.93& 4.71& 201.6& 0.79 & 0.56  \\
$\gamma = 1.0$ & 15.64 & 5.31 & 90.5 & 0.63 & 0.64 & 3.93 & 4.74 & 201.8 & 0.78 & 0.55  \\
\bottomrule
\end{tabular*}
\end{table*}

\section{More Results for Improvement by MixFlow Training (Section~\ref{sec:improvementbymixflow})}
\noindent\textbf{Results for SDE sampling.}
We present
the results for SDE samplers~\cite{ma2024sit} in \Cref{tab:sitflowmatchingSDEsolver}. 
The evaluation settings are the same: same guidance scale ($1.5$), and same sampling steps ($250$). 
The sampling scheme is first-order Euler-Maruyama integrator. 
The observations are the same
for the ODE sampler.

\vspace{1mm}
\noindent\textbf{Input Perturbation~\cite{ning2023input}.}
Input Perturbation trains the diffusion model
by
conducting a Gaussian perturbation
on the ground truth noisy data
to simulate the inference time prediction errors.
We implement the input perturbation algorithm 
by carefully tuning
the perturbation strength.

\begin{table*}[!th]
\footnotesize
\setlength{\tabcolsep}{1pt} 
\caption{\textbf{The effect of perturbation strength in DDPM-IP.}
The performance is sensitive to perturbation strength.} 
\label{tab:ddpmipstudy}
\begin{tabular*}{\textwidth}{@{} l  @{\extracolsep{\fill}}  c c c c c  c c c c c@{}}
\toprule
  \multirow{2}{*}{\textbf{ Strength }} &  \multicolumn{5}{c}{\textbf{Generation  w/o guidance}} & \multicolumn{5}{c}{\textbf{Generation  w/ guidance}} \\
 \cmidrule(lr){2-6} \cmidrule(lr){7-11}
    & gFID $\downarrow$ & sFID$\downarrow$ & IS $\uparrow$ & Pre. $\uparrow$ & Rec. $\uparrow$ & gFID $\downarrow$ & sFID $\downarrow$ & IS $\uparrow$ & Pre. $\uparrow$ & Rec. $\uparrow$ \\
\midrule

$0.1$ & 17.26 & 5.82 & 81.8 & 0.62 & 0.66 & 4.36 & 4.73 & 183.9 & 0.78 & 0.57 \\
$0.15$ & 17.01 & 5.37 & 83.1 & 0.62 & 0.65 & 4.32 & 4.77 & 185.8 & 0.78 & 0.56 \\
$0.2$ & 17.55 & 6.44 & 81.1 & 0.62 & 0.65 & 4.86 & 6.56  & 180.3 & 0.78 & 0.55 \\
$0.3$ & 17.82 & 8.45 & 80.2 & 0.62 & 0.62 & 5.41 & 9.08  & 178.9 & 0.78 & 0.53 \\
$0.5$ & 33.50 & 61.8 & 51.5 & 0.50 & 0.55 & 14.46 & 47.4 & 125.4 & 0.68 & 0.44 \\

\bottomrule
\end{tabular*}
\end{table*}

\Cref{tab:ddpmipstudy} studies the effect of perturbation strengths.
The performance is sensitive to the perturbation strength:
too large or too small lead to poor result.
In the main paper,
we report the best one
for strength $0.15$.

\vspace{1mm}
\noindent\textbf{Time Shift sampler~\cite{li2023timeshift}.}
Time-Shift Sampler modifies the sampling process by adjusting the timestep for the next sampling iteration based on the previously sampled data.
\Cref{tab:timeshift} presents 
the results on how the two hyperparameters, 
window size and cutoff value,
affect the performance.
We report the best result
in the main paper.

\begin{table*}[!th]
\footnotesize
\setlength{\tabcolsep}{1pt} 
\caption{\textbf{The effect of scaling strength $\lambda(t) = kt + b$ in Epsilon Scaling.}
The performance is sensitive to scaling strength.} 
\label{tab:epsilonscaling}
\begin{tabular*}{\textwidth}{@{} l  @{\extracolsep{\fill}}  c c c c c  c c c c c@{}}
\toprule
 &  \multicolumn{5}{c}{\textbf{Generation  w/o guidance}} & \multicolumn{5}{c}{\textbf{Generation  w/ guidance}} \\
 \cmidrule(lr){2-6} \cmidrule(lr){7-11}
    & gFID $\downarrow$ & sFID$\downarrow$ & IS $\uparrow$ & Pre. $\uparrow$ & Rec. $\uparrow$ & gFID $\downarrow$ & sFID $\downarrow$ & IS $\uparrow$ & Pre. $\uparrow$ & Rec. $\uparrow$ \\
\midrule

$k=0, b = 1.02$ & 18.13 & 6.55 & 78.1 & 0.60 & 0.65 & 4.78 & 5.12 & 179.1 & 0.77 & 0.57\\
$k=0, b = 1.015$ & 17.87 & 6.41 & 79.1 & 0.62 & 0.66 & 4.49 & 4.94 & 181.6 & 0.78 & 0.56\\
$k=0, b = 1.01$ & 17.47 & 6.27 & 79.6 & 0.62 & 0.66 & 4.46 & 4.90 & 182.0 & 0.78 & 0.56\\
$k=0, b = 1.005$ & 17.28 & 6.28 & 80.3 & 0.62 & 0.66 & 4.40 & 4.80 & 182.4 & 0.78 &0.56\\
$k=0, b = 0.995$ & 18.19 & 6.55 & 77.9 & 0.61 & 0.65& 4.87 & 5.22 & 178.1 & 0.77 & 0.57 \\
$k=0, b = 0.99$ & 18.87 & 6.87 & 75.01 & 0.60 & 0.65 & 5.07 & 5.43 & 175.1 & 0.77 & 0.57\\
$k=0.0001, b = 1.005$ & 17.18 & 6.28 & 80.8 & 0.62 & 0.65 & 4.39 & 4.81 & 182.5 & 0.78 & 0.56\\
$k=0.0002, b = 1.005$ & 17.25 & 6.31 & 80.1 & 0.62 & 0.65 & 4.42 & 4.91 & 182.1 & 0.78 & 0.56\\
$k=0.0004, b = 1.005$ & 17.51 & 6.30 & 79.1 & 0.62 & 0.64 & 4.47 & 4.97 & 179.1 & 0.78 & 0.56\\
\bottomrule
\end{tabular*}
\end{table*}

\vspace{1mm}
\noindent\textbf{Epsilon Scaling sampler~\cite{ning2023epsilonscaling}.}
Epsilon Scaling adjusts the sampling process
by scaling the noise prediction, mitigating the input mismatch between training and sampling.
It moves the sampling trajectory
closer to the vector field learned in the training phase by scaling the network output epsilon.

\Cref{tab:epsilonscaling} studies the effect of scaling strength $\lambda(t) = kt + b$.
The performance is sensitive to the choice of $(k, b)$.
We report the best result in the main paper.

\vspace{1mm}
\noindent\textbf{Text-to-image generation.}
We present the results for the sub-metrics from GenEval and T2I-CompBench.
\Cref{tab:t2icompbench} shows the results for T2I-CompBench across six attributes: Color, Shape, Texture, Spatial, Non-Spatial, and Complex.
\Cref{tab:t2igeneval} shows the results for GenEval across six categories: Single Object, Two Objects, Counting, Colors, Position, and Attribute Binding.
MixFlow consistently improves the performance across all the metrics.

The high-quality T2I dataset
contains $4.5$M samples
from public sources
(Laion-5B, CC12M, Journey-DB and ShareGPT4V)
with quality improved by filtering
and re-captioning with Qwen2.5-VL-7B-Instruct
and Blip3o,
and $0.5$M synthetic $1024 \times 1024$ images
generated by SD3.5 and FLUX.

\begin{figure*}[t]
  \centering
  \footnotesize 
\subfloat[Standard Training]{\includegraphics[width=0.23\linewidth]{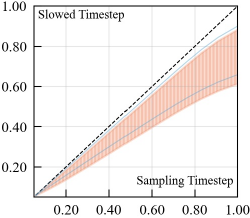}}
~~~~~~
\subfloat[MixFlow Training, $\gamma = 0.4$]{\includegraphics[width=0.23\linewidth]{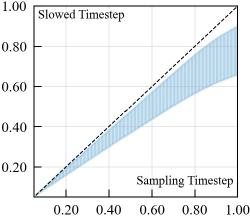}}
~~~~~~
\subfloat[MixFlow Training with $\gamma=0.8$]{\includegraphics[width=0.23\linewidth]{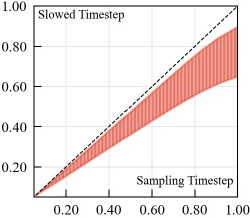}}
~~~~~~
\subfloat[Upper and lower envelopes of $\gamma = 0.8$ and $\gamma = 0.4$]{\includegraphics[width=0.23\linewidth]{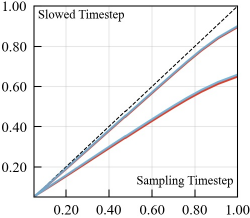}}
  \caption{
   \textbf{Slow Flow for the RAE model.}}
 \label{fig:slowflow_rae}
\end{figure*}

\vspace{1mm}
\noindent\textbf{RAE.}
\Cref{fig:slowflow_rae}
presents the slow flow phenomenon 
that is observed in the RAE models.
The overall observation is similar:
MixFlow training makes slowed timesteps
be closer to sampling timesteps,
and the range for slowed timesteps be smaller. 
One difference from Figure~\ref{fig:slowflow} in the main paper
and \Cref{fig:slowflow_gvp}
is that
the minimum slowed timestep is larger.
For example,
in \Cref{fig:slowflow_rae} (a),
the minimum slowed timestep around sampling timestep $1.00$
is about $0.6$,
larger than $0.2$ in Figure~\ref{fig:slowflow} 
in the main paper.
This implies that the slowed timestep 
for RAE is possible to be closer
to the sampling timestep.

Regarding  the mixture range coefficient
$\gamma$,
we tried two choices:
$0.8$ that is selected 
for all the other experiments,
and $0.4$ ($=1.00-0.60$)
that is based on the observation
in \Cref{fig:slowflow_rae}.
Both achieve similar performance
(See \Cref{tab:RAE_gamma} and \Cref{fig:slowflow_rae}).
$\gamma = 0.8$
needs $430$ training epochs
and $\gamma = 0.4$
needs around a half,
$200$,
training epochs.
\Cref{fig:slowflow_rae} (d)
shows that 
the upper and lower envelopes of slowed timesteps for $\gamma = 0.4$
and $\gamma = 0.8$
are almost the same.
In the main paper,
we report the results for 
$\gamma = 0.4$.

We additionally report the performance of RAE with $20$ sampling steps in \Cref{tab:RAE}.
One can see that 
the improvement from MixFlow is larger than that for $50$ sampling steps.

\begin{table*}[!th]
\footnotesize
\setlength{\tabcolsep}{1pt} 
\caption{\textbf{The effect of window size and cutoff value  in Time-Shift Sampler.}}
\label{tab:timeshift}
\begin{tabular*}{\textwidth}{@{} l  @{\extracolsep{\fill}}  c c c c c  c c c c c@{}}
\toprule
 &  \multicolumn{5}{c}{\textbf{Generation  w/o guidance}} & \multicolumn{5}{c}{\textbf{Generation  w/ guidance}} \\
 \cmidrule(lr){2-6} \cmidrule(lr){7-11}
    & gFID $\downarrow$ & sFID$\downarrow$ & IS $\uparrow$ & Pre. $\uparrow$ & Rec. $\uparrow$ & gFID $\downarrow$ & sFID $\downarrow$ & IS $\uparrow$ & Pre. $\uparrow$ & Rec. $\uparrow$ \\
\midrule

$w=5, \text{cutoff} =500$ & 18.24 & 6.57 & 78.1 & 0.61 & 0.66 & 4.54 & 4.89 & 177.1 & 0.77 & 0.56\\
$w=5, \text{cutoff} =400$ & 18.20 & 6.50 & 78.3 & 0.61 & 0.66 & 4.54 & 4.89 & 178.4 & 0.77 & 0.56\\
$w=5, \text{cutoff} =300$ &18.17 & 6.48 & 79.1 & 0.62 & 0.66 & 4.52 & 4.87 & 179.2 & 0.78 & 0.56\\
$w=5, \text{cutoff} =200$ & 18.14 & 6.44 & 79.5 & 0.62 & 0.66 & 4.50 & 4.82 & 180.9 & 0.78 & 0.56\\
$w=5, \text{cutoff} =100$ &18.13 & 6.43 & 79.9 & 0.62 & 0.66 & 4.49 & 4.82 & 181.7 & 0.78 & 0.56\\
$w=5, \text{cutoff} =50$ & 18.15 & 6.45 & 79.1 & 0.62 & 0.66 & 4.50 & 4.82 & 180.9 & 0.78 & 0.56\\
$w=2, \text{cutoff} =100$ &18.13 & 6.40 & 80.1 & 0.62 & 0.66 & 4.49 & 4.90 & 181.2 & 0.78 & 0.56\\
$w=8, \text{cutoff} =100$ & 18.12 & 6.41 & 79.5 & 0.62 & 0.66 & 4.49 & 4.80 & 180.5 & 0.78 & 0.56\\
$w=10, \text{cutoff} =100$ & 18.14 & 6.44 & 79.1 & 0.62 & 0.66 & 4.50 & 4.80 & 180.3 & 0.78 & 0.56\\
\bottomrule
\end{tabular*}
\end{table*}

\begin{table}[t]
\centering
    \footnotesize   
\setlength{\tabcolsep}{1pt} 
\caption{\textbf{The performance for text-to-image generation in T2I-CompBench.} 
}
\label{tab:t2icompbench}
\begin{tabular}{lcccccc}
\toprule
\textbf{Method}  &  Color $\uparrow$ & Shape  $\uparrow$& Texture $\uparrow$  & Spatial $\uparrow$ & Non-Spatial $\uparrow$ & Complex $\uparrow$\\
   \midrule
  SD $3.5$   & 0.8135 & 0.6555 & 0.7841 & 0.2850 & 0.3129 & 0.4202 \\
     SD $3.5$-ft-20k& 0.8142 & 0.6556 & 0.7842 & 0.2852 & 0.3133 & 0.4208\\
   SD $3.5$-ft-10k&  0.8140 & 0.6556 & 0.7839 & 0.2855 & 0.3132 & 0.4205\\
  + MixFlow  & 0.8612 & 0.7013 & 0.8112 & 0.3642 & 0.3402 & 0.4756\\
\bottomrule
\end{tabular}
\end{table}    

\begin{table}
    \footnotesize 
        \centering
        \caption{\textbf{The performance for text-to-image generation in Geneval.}}
        \label{tab:t2igeneval}
        \setlength{\tabcolsep}{1.5pt} 
        \begin{tabular*}{\linewidth}{@{\extracolsep{\fill}} l cccccc}
            \toprule
            \multirow{2}{*}{\textbf{Method}} & Single & Two & Count- & \multirow{2}{*}{Colors} & Posi- & Attr. \\
             & Obj. & Obj. & ing & & tion & Bind. \\
            \midrule
            SD $3.5$ & 0.98 & 0.78 & 0.50 & 0.81 & 0.24 & 0.52 \\
            SD $3.5$-ft-20k & 0.98 & 0.78 & 0.51 & 0.80 & 0.24 & 0.52 \\
            SD $3.5$-ft-10k & 0.98 & 0.78 & 0.51 & 0.80 & 0.24 & 0.51 \\
            + MixFlow & 0.98 & 0.79 & 0.63 & 0.82 & 0.27 & 0.56 \\
            \bottomrule
        \end{tabular*}
    \end{table}

\begin{table*}[t!]
    \centering
    \footnotesize
    \setlength{\tabcolsep}{2pt}
    \caption{\textbf{The effect of mixture range coefficient $\gamma$ for RAE.}}
    \label{tab:RAE_gamma}
    \begin{tabular*}{\linewidth}{@{\extracolsep{\fill}} lc ccccc ccccc}
        \toprule
        \multirow{2}{*}{\textbf{$\gamma$}} & \multirow{2}{*}{\textbf{Epochs}} & \multicolumn{5}{c}{\textbf{Generation w/o guidance}} & \multicolumn{5}{c}{\textbf{Generation w/ guidance}} \\
        \cmidrule(lr){3-7} \cmidrule(lr){8-12}
         & & gFID $\downarrow$ & sFID$\downarrow$ & IS $\uparrow$ & Pre. $\uparrow$ & Rec. $\uparrow$ & gFID $\downarrow$ & sFID $\downarrow$ & IS $\uparrow$ & Pre. $\uparrow$ & Rec. $\uparrow$ \\
        \midrule
        $0.4$ & 200 & 1.43 & 4.81 & 239.8 & 0.80 & 0.64 & 1.10 & 4.40 & 259.7 & 0.78 & 0.67 \\
        \arrayrulecolor{black!30}\midrule 
        $0.8$ & 200 & 1.57 & 4.95 & 233.5 & 0.80 & 0.63 & 1.21 & 4.60 & 248.9 & 0.78 & 0.65 \\
        $0.8$ & 430 & 1.43 & 4.85 & 239.1 & 0.80 & 0.64 & 1.10 & 4.42 & 259.3 & 0.78 & 0.66 \\
        \arrayrulecolor{black}\bottomrule 
    \end{tabular*}
\end{table*}
   
\begin{table*}[t]
\footnotesize
\setlength{\tabcolsep}{1pt} 
\caption{\textbf{MixFlow on RAE.}
The gain for $20$ sampling steps
is larger than $50$ sampling steps.} 
\label{tab:RAE}
\begin{tabular*}{\textwidth}{@{} l @{\extracolsep{\fill}}  c c c c c  c c c c c@{}}
\toprule
  \multirow{2}{*}{\textbf{ Method }}   & \multicolumn{5}{c}{\textbf{Generation  w/o guidance}} & \multicolumn{5}{c}{\textbf{Generation  w/ guidance}} \\
\cmidrule(lr){2-6} \cmidrule(lr){7-11}
 &  gFID $\downarrow$ & sFID$\downarrow$ & IS $\uparrow$ & Pre. $\uparrow$ & Rec. $\uparrow$ & gFID $\downarrow$ & sFID $\downarrow$ & IS $\uparrow$ & Pre. $\uparrow$ & Rec. $\uparrow$ \\
\midrule
\multicolumn{11}{l}{\emph{$50$ sampling steps}}\\ 
RAE & 1.51 & 5.31 & 242.9 & 0.79 & 0.63 & 1.13 & 4.74 & 262.6 & 0.78 & 0.67 \\
+ MixFlow  &  1.43 & 4.81 & 239.8 & 0.80 & 0.64 & 1.10 & 4.40 & 259.7 & 	0.78	& 0.67 \\
\midrule

\multicolumn{11}{l}{\emph{$20$ sampling steps}}\\ 
RAE &  1.92 &	6.32&	232.8&	0.78&	0.64 &	1.35	&5.37&	254.4&	0.77&	0.67\\

+ MixFlow   &  1.64 & 5.39 & 238.7 & 0.80 & 0.63 & 1.18 & 4.77 &255.8 & 0.78 & 0.66  \\
\bottomrule
\end{tabular*}
\end{table*}

    \begin{table}
    \footnotesize
        \centering
        \caption{\textbf{Class-conditional performance on ImageNet $512 \times 512$} without guidance.}
        \label{tab:appendixsotaresults512}
        \setlength{\tabcolsep}{7pt}
        \begin{tabular}{lrrrrr}
            \toprule
            \textbf{Method} & gFID$\downarrow$ & sFID$\downarrow$ & IS$\uparrow$ & Prec.$\uparrow$ & Rec.$\uparrow$ \\
            \midrule
            \multicolumn{6}{@{}l}{\textit{Autogressive}} \\
            \arrayrulecolor{black!30}\midrule
            MAGViT-v2~\cite{yu2023language} & 3.07 & -- & 213.1 & -- & -- \\
            \arrayrulecolor{black}\midrule
            \multicolumn{6}{@{}l}{\textit{Diffusion}} \\
            \arrayrulecolor{black!30}\midrule
            DiT~\cite{peebles2023DiT} & 11.93 & -- & -- & -- & -- \\
            SiT~\cite{ma2024sit} & 9.20 & 7.33 & 125.3 & 0.78 & \textbf{0.64} \\
            EDM2~\cite{karras2024analyzing} & 1.91 & -- & -- & -- & -- \\
            \rowcolor{black!5}
            RAE~\cite{zheng2025rae} & 1.61 & 4.96 & 234.1 & 0.81 & 0.63 \\
            \arrayrulecolor{black}\midrule
            \rowcolor{black!5} MixFlow + RAE & \textbf{1.55} & \textbf{4.67} & \textbf{237.0} & \textbf{0.82} & 0.60 \\
            \arrayrulecolor{black}\bottomrule
        \end{tabular}
\end{table}
\section{SOTA Results on ImageNet $512 \times 512$
without Guidance.}
\label{appendix:sotawoguidance512}

We present the comparison with state-of-the-art models on ImageNet $512 \times 512$ for generation without guidance in \Cref{tab:appendixsotaresults512}. 
We compare with the methods that have reported performance on ImageNet $512 \times 512$ without guidance.
We follow the same evaluation setting as in the main paper.
MixFlow + RAE achieves the best performance across multiple metrics,
including gFID, sFID, IS and Precision.

\section{Improvement over Qwen-Image}
\label{appendix:qwenimage}
We verify MixFlow
on Qwen-Image (20B MMDiT),
that is larger than SD 3.5
and differs in text encoder and RoPEs.
The setup follows the same procedure as SD 3.5:
we finetune the model for $10$K steps
and then post-train with MixFlow for $10$K steps.
A baseline model (ft-$20$K)
is trained for $20$K steps
without MixFlow.

\Cref{tab:qwenimage} presents the results.
MixFlow consistently improves the performance
across both GenEval and DPG-Bench
for both $50$ and $10$ sampling steps.
This indicates that MixFlow
is effective for larger models
with different architectures.

\begin{table}[t]
\centering
    \footnotesize
    \setlength{\tabcolsep}{7pt}
\caption{\textbf{The performance of MixFlow over Qwen-Image.}
}
\label{tab:qwenimage}
\begin{tabular}{lcccc}
\toprule
\multirow{2}{*}{\textbf{Method}} & \multicolumn{2}{c}{GenEval Avg. $\uparrow$} & \multicolumn{2}{c}{DPG-Bench Avg. $\uparrow$} \\
\cmidrule(lr){2-3} \cmidrule(lr){4-5}
 & 50 steps & 10 steps & 50 steps & 10 steps \\
\midrule
Qwen-Image & 0.87 & 0.75 & 88.38 & 81.47 \\
Qwen-Image-ft-20k & 0.87 & 0.75 & 88.46 & 81.54 \\
Qwen-Image-ft-10k & 0.87 & 0.75 & 88.50 & 81.50 \\
+ MixFlow & \textbf{0.90} & \textbf{0.86} & \textbf{90.32} & \textbf{87.65} \\
\bottomrule
\end{tabular}
\end{table}

\section{Qualitative  Results}
\label{appendix:visualresultsimagenet}
\begin{figure}[t]
  \centering
  \footnotesize
    \centering
{\includegraphics[width=1.0\linewidth]{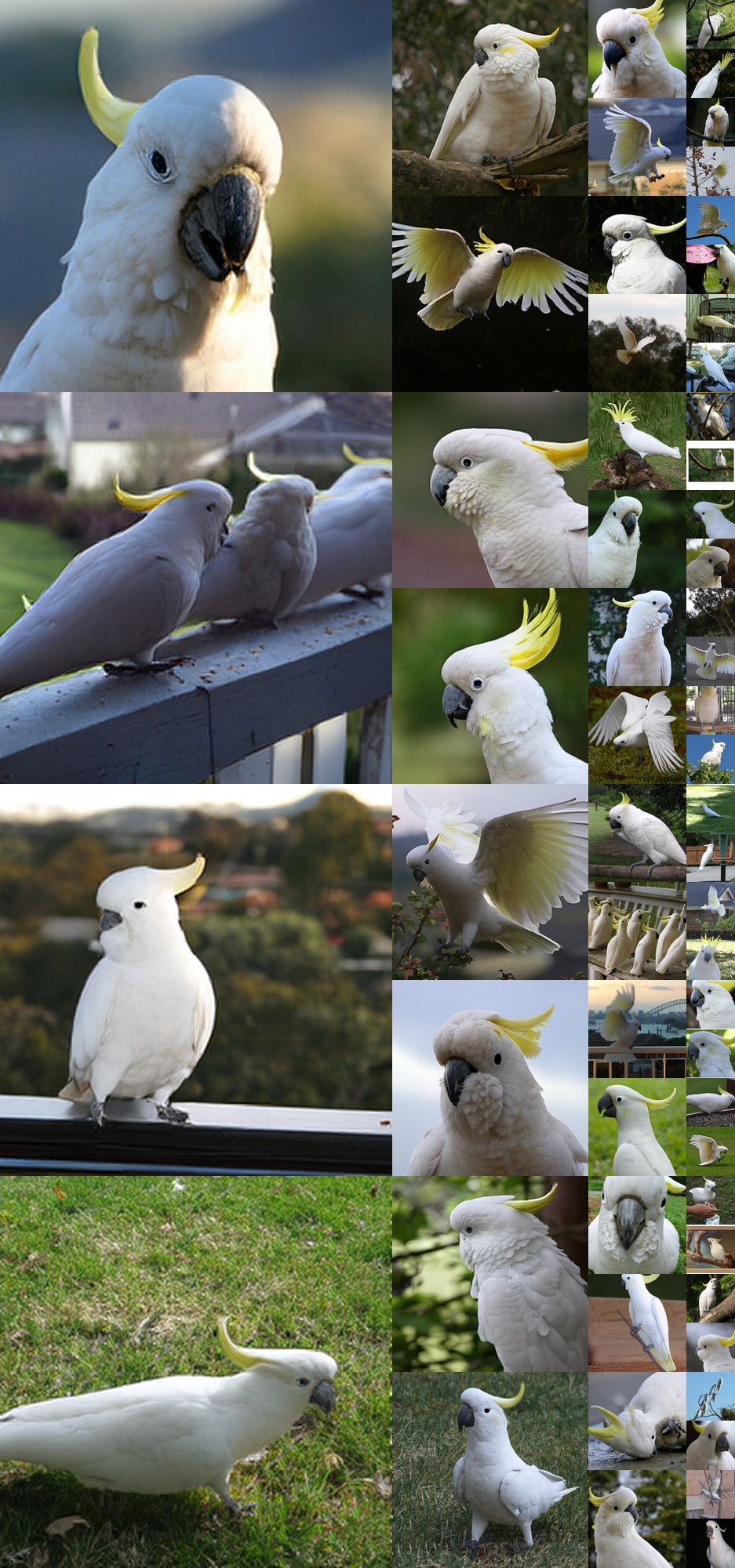}}
  \caption{
   \textbf{Uncurated $256 \times 256$ MixFlow-XL samples.}  \\
   AutoGuidance Scale = 1.5\\
   Class label =  ``Sulphur-crested cockatoo'' (89)}
 \label{fig:256_089}
\end{figure}

\begin{figure}[t]
  \centering
  \footnotesize
    \centering
{\includegraphics[width=1.0\linewidth]{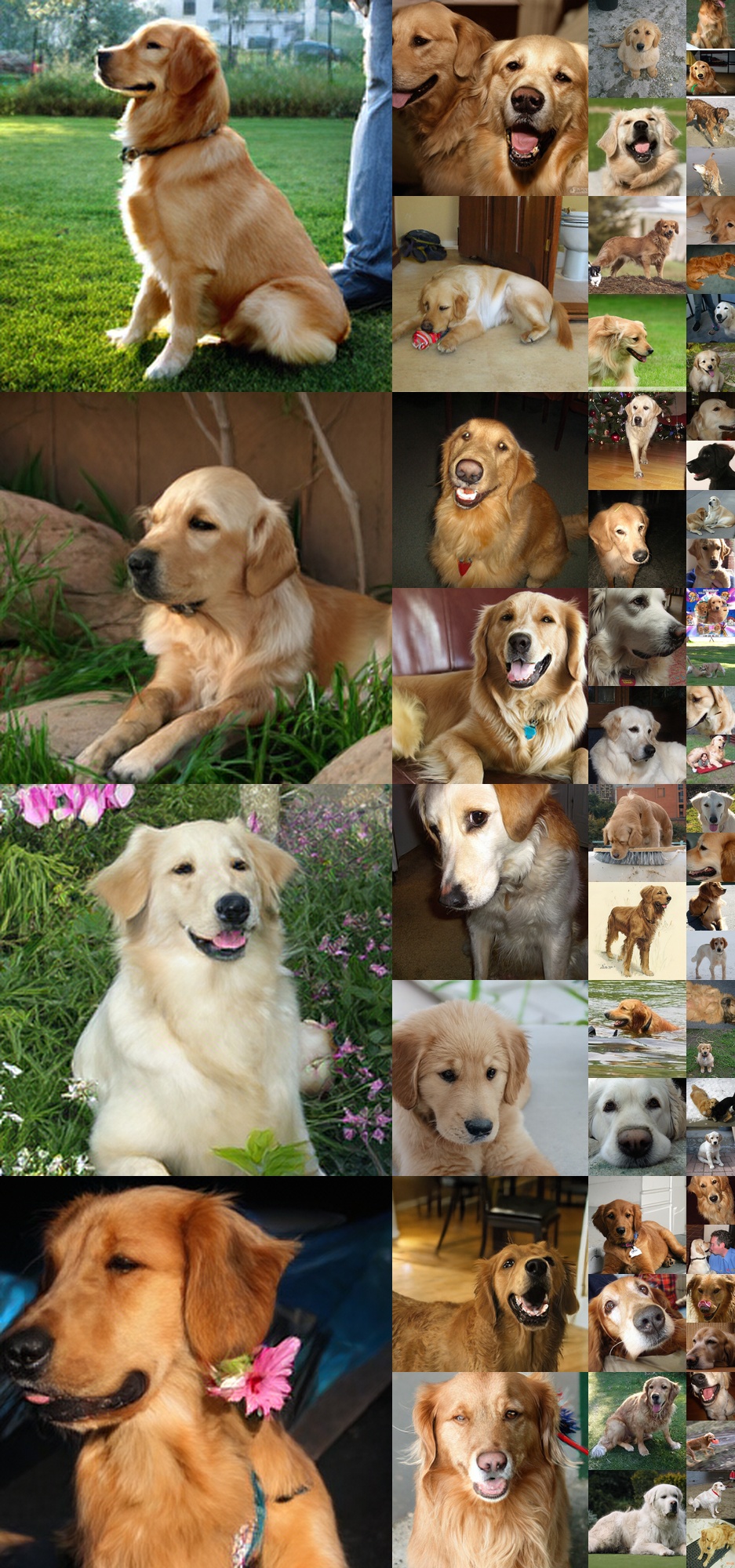}}
  \caption{
   \textbf{Uncurated $256 \times 256$ MixFlow-XL samples.}  \\
   AutoGuidance Scale = 1.5\\
   Class label = ``golden retriever'' (207)}
 \label{fig:256_207}
\end{figure}

\begin{figure}[t]
  \centering
  \footnotesize
    \centering
{\includegraphics[width=1.0\linewidth]{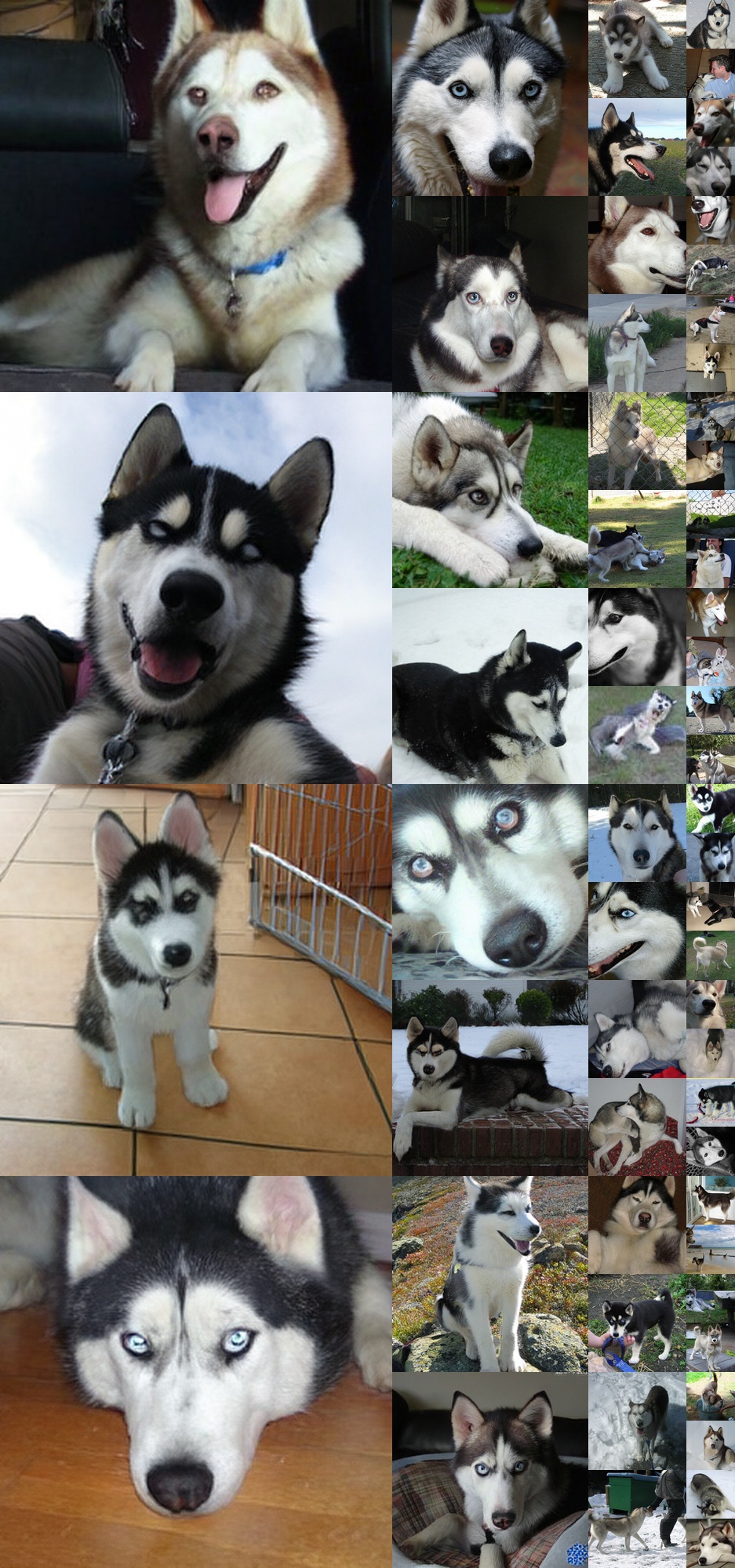}}
  \caption{
   \textbf{Uncurated $256 \times 256$ MixFlow-XL samples.}  \\
   AutoGuidance Scale = 1.5\\
   Class label = ``husky'' (250)}
 \label{fig:256_250}
\end{figure}

\begin{figure}[t]
  \centering
  \footnotesize
    \centering
{\includegraphics[width=1.0\linewidth]{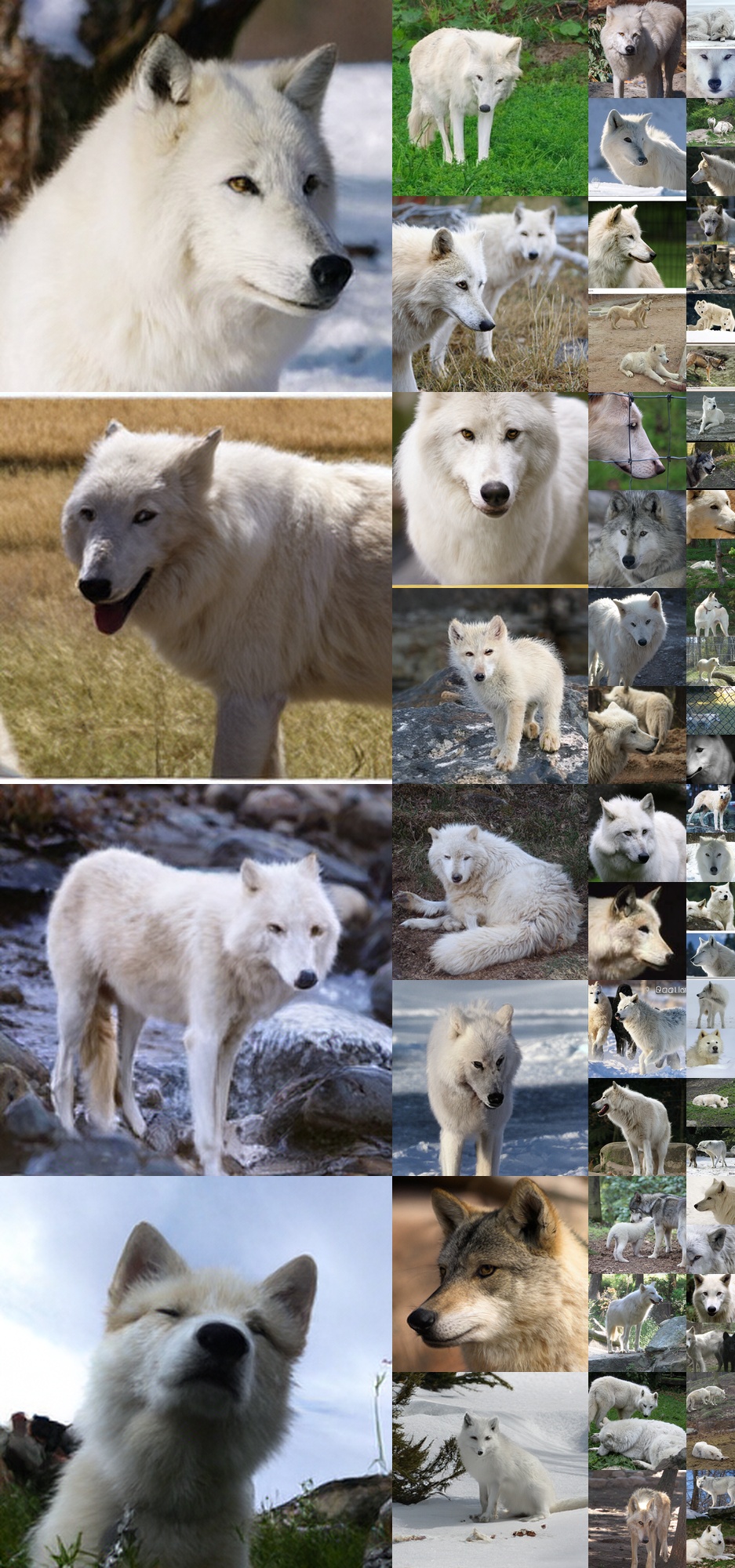}}
  \caption{
   \textbf{Uncurated $256 \times 256$ MixFlow-XL samples.}  \\
   AutoGuidance Scale = 1.5\\
   Class label = ``arctic wolf'' (270)}
 \label{fig:256_270}
\end{figure}

\begin{figure}[t]
  \centering
  \footnotesize
    \centering
{\includegraphics[width=1.0\linewidth]{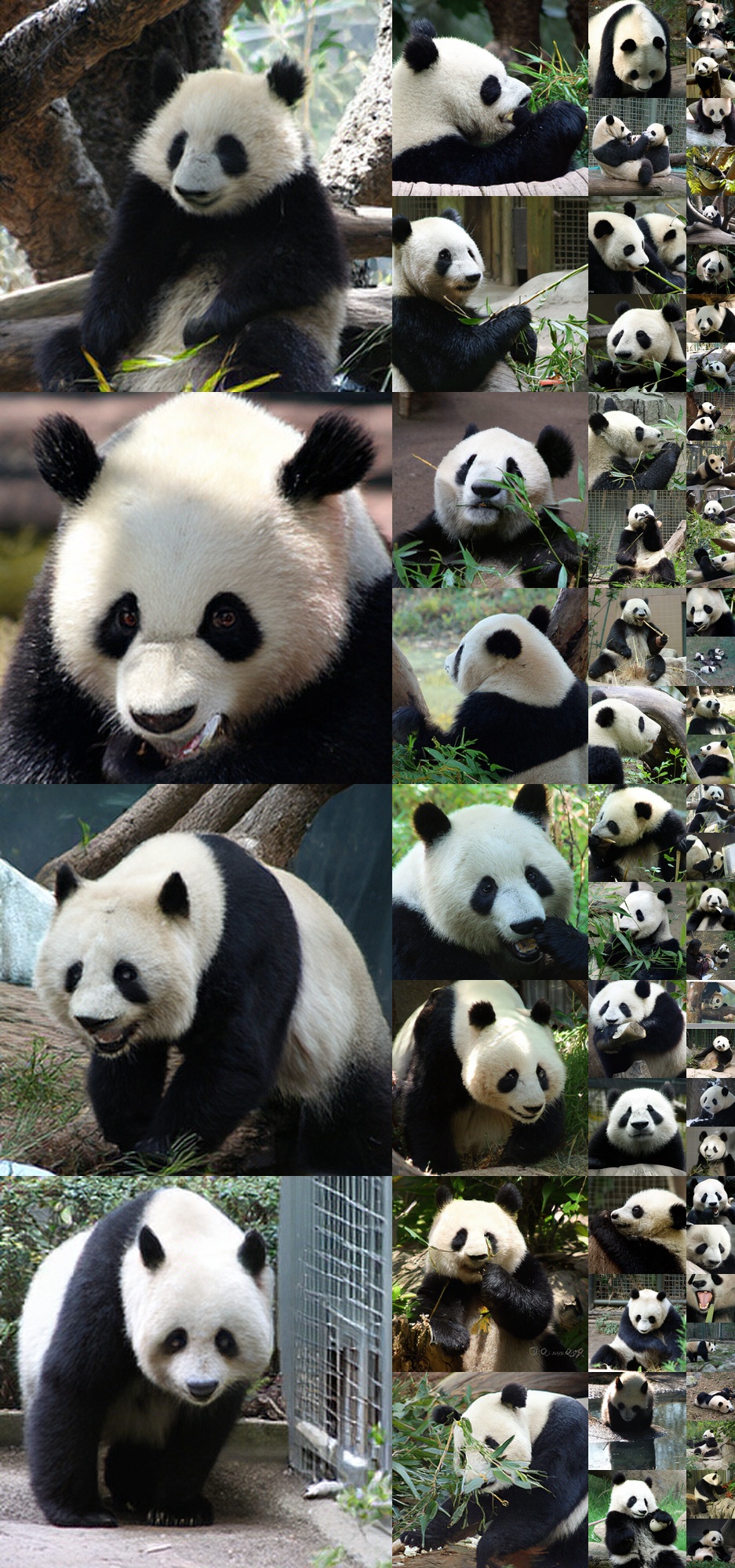}}
  \caption{
   \textbf{Uncurated $256 \times 256$ MixFlow-XL samples.}  \\
   AutoGuidance Scale = 1.5\\
   Class label = ``panda'' (388)}
 \label{fig:256_388}
\end{figure}

\begin{figure}[t]
  \centering
  \footnotesize
    \centering
{\includegraphics[width=1.0\linewidth]{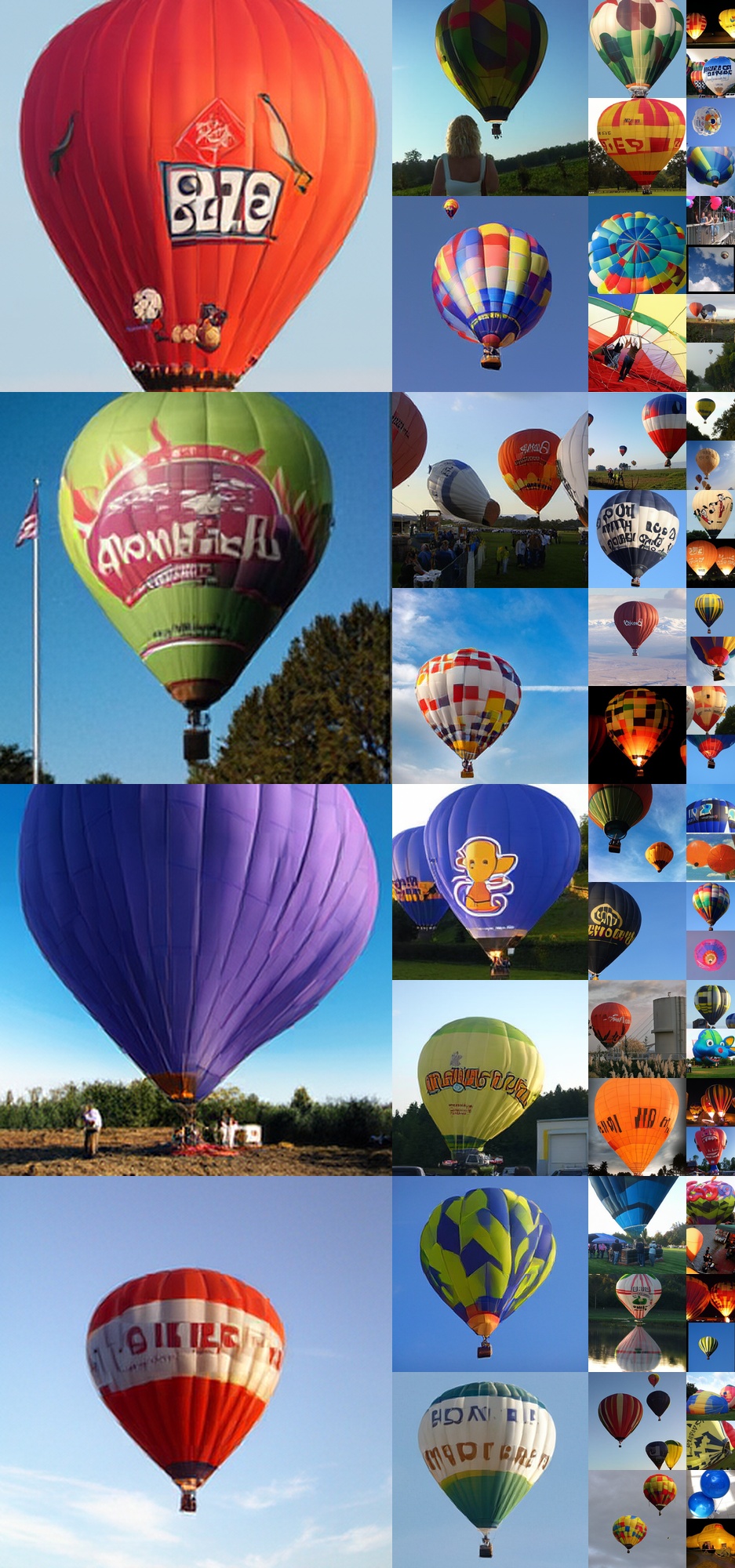}}
  \caption{
   \textbf{Uncurated $256 \times 256$ MixFlow-XL samples.}  \\
   AutoGuidance Scale = 1.5\\
   Class label = ``balloon'' (417)}
 \label{fig:256_417}
\end{figure}

\begin{figure}[t]
  \centering
  \footnotesize
    \centering
{\includegraphics[width=1.0\linewidth]{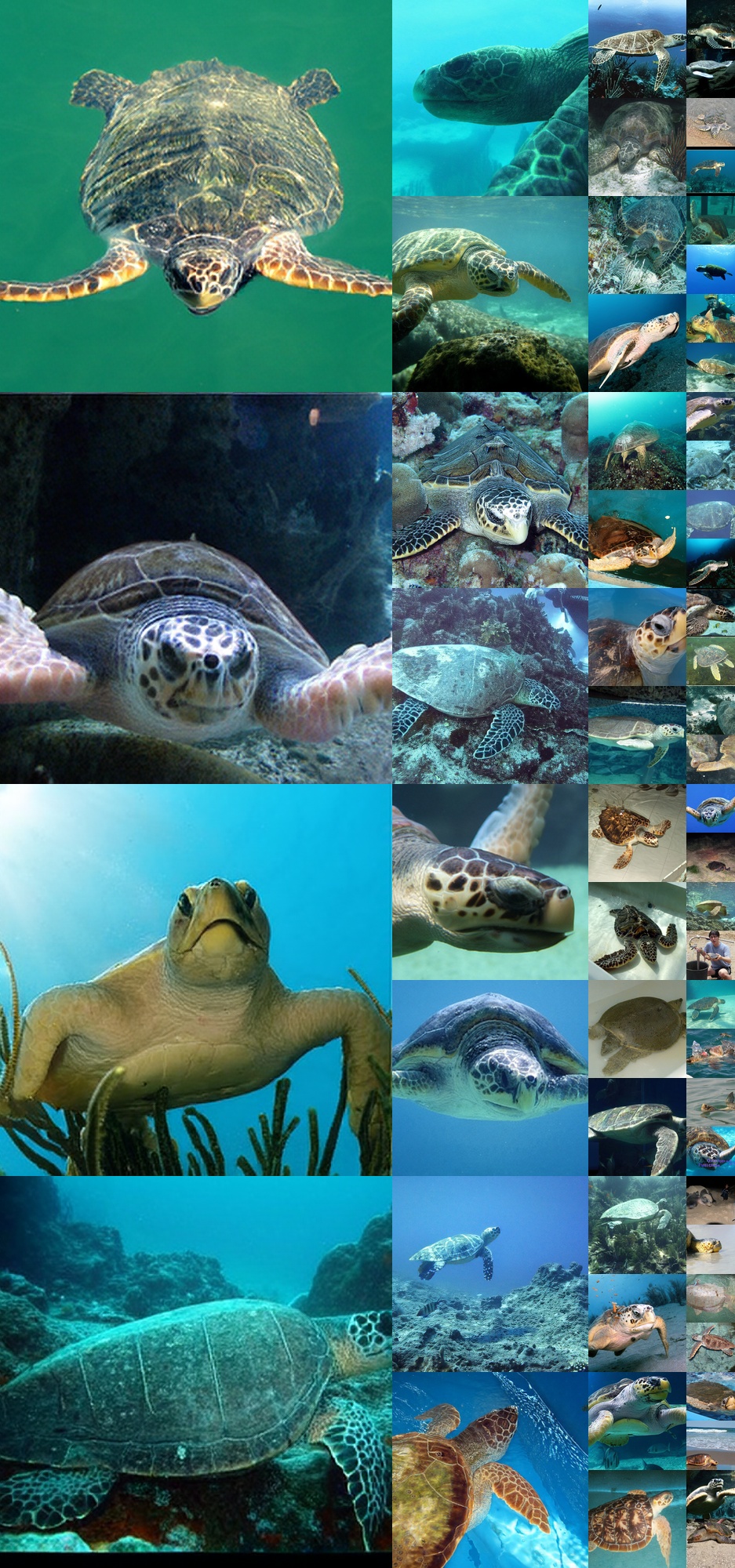}}
  \caption{
   \textbf{Uncurated $512 \times 512$ MixFlow-XL samples.}  \\
   AutoGuidance Scale = 1.5\\
   Class label = ``Loggerhead sea turtle'' (33)}
 \label{fig:512_33}
\end{figure}

\begin{figure}[t]
  \centering
  \footnotesize
    \centering
{\includegraphics[width=1.0\linewidth]{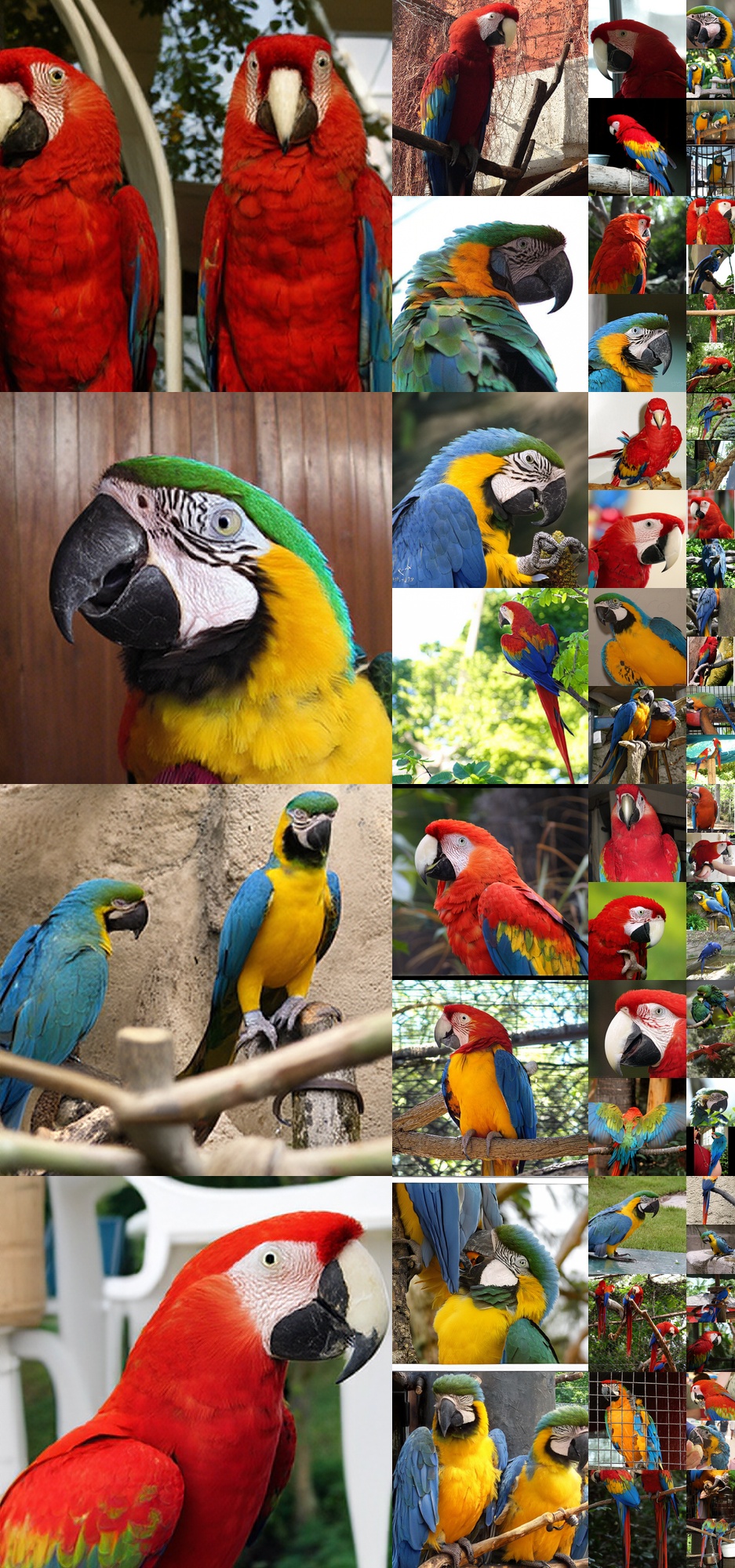}}
  \caption{
   \textbf{Uncurated $512 \times 512$ MixFlow-XL samples.}  \\
   AutoGuidance Scale = 1.5\\
   Class label = ``macaw'' (88)}
 \label{fig:512_88}
\end{figure}

\begin{figure}[t]
  \centering
  \footnotesize
    \centering
{\includegraphics[width=1.0\linewidth]{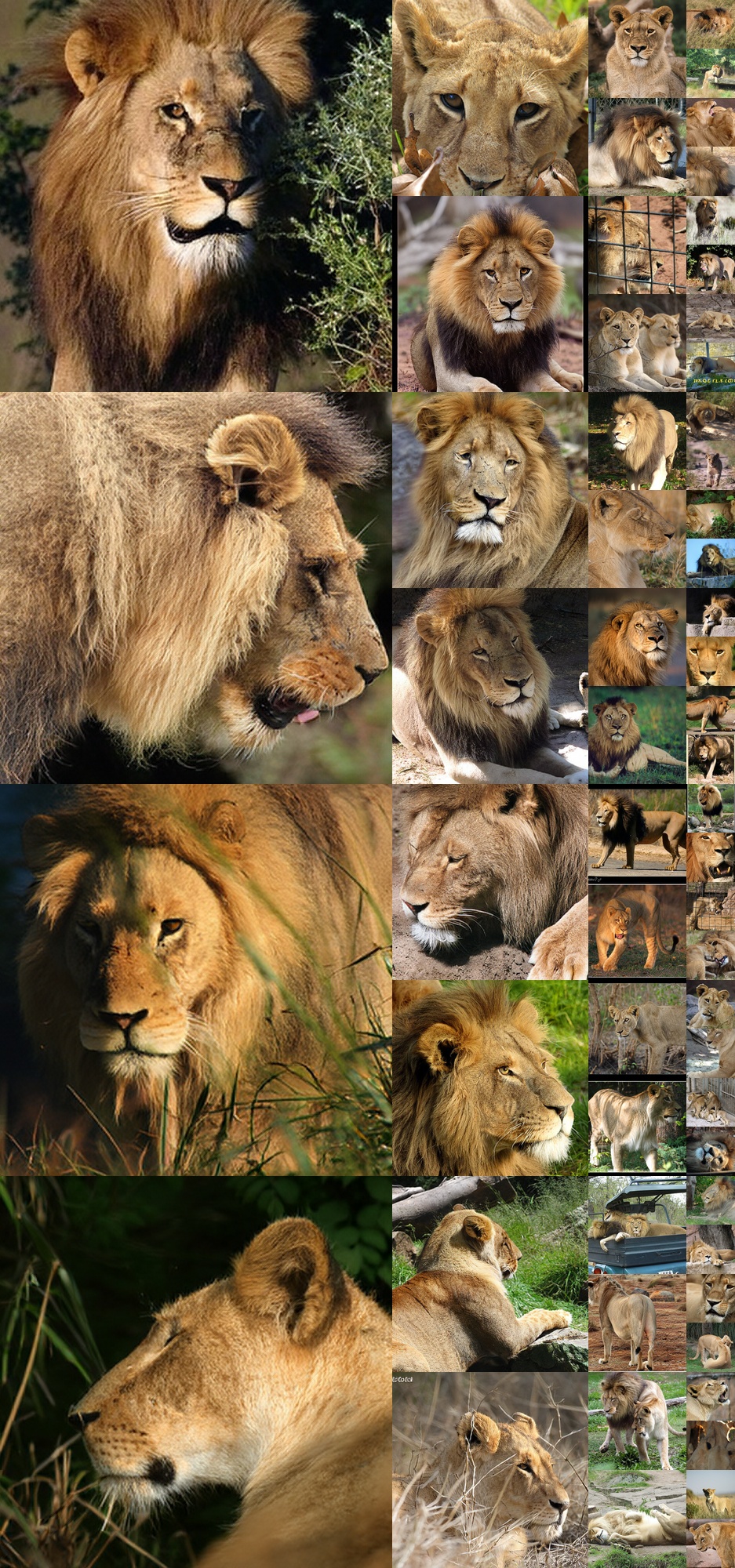}}
  \caption{
   \textbf{Uncurated $512 \times 512$ MixFlow-XL samples.}  \\
   AutoGuidance Scale = 1.5\\
   Class label = ``lion'' (291)}
 \label{fig:512_291}
\end{figure}

\begin{figure}[t]
  \centering
  \footnotesize
    \centering
{\includegraphics[width=1.0\linewidth]{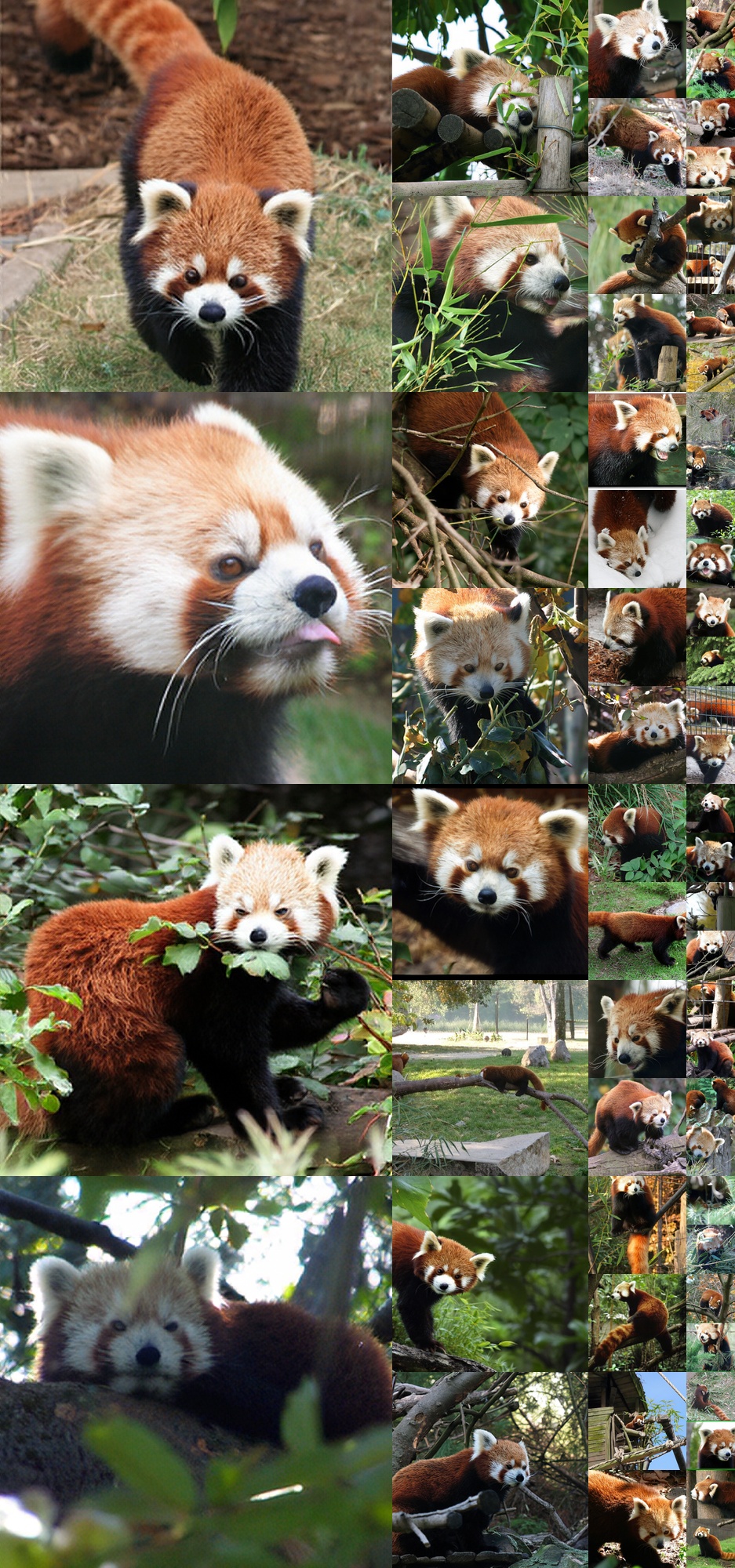}}
  \caption{
   \textbf{Uncurated $512 \times 512$ MixFlow-XL samples.}  \\
   AutoGuidance Scale = 1.5\\
   Class label = ``red panda'' (387)}
 \label{fig:512_387}
\end{figure}

\begin{figure}[t]
  \centering
  \footnotesize
    \centering
{\includegraphics[width=1.0\linewidth]{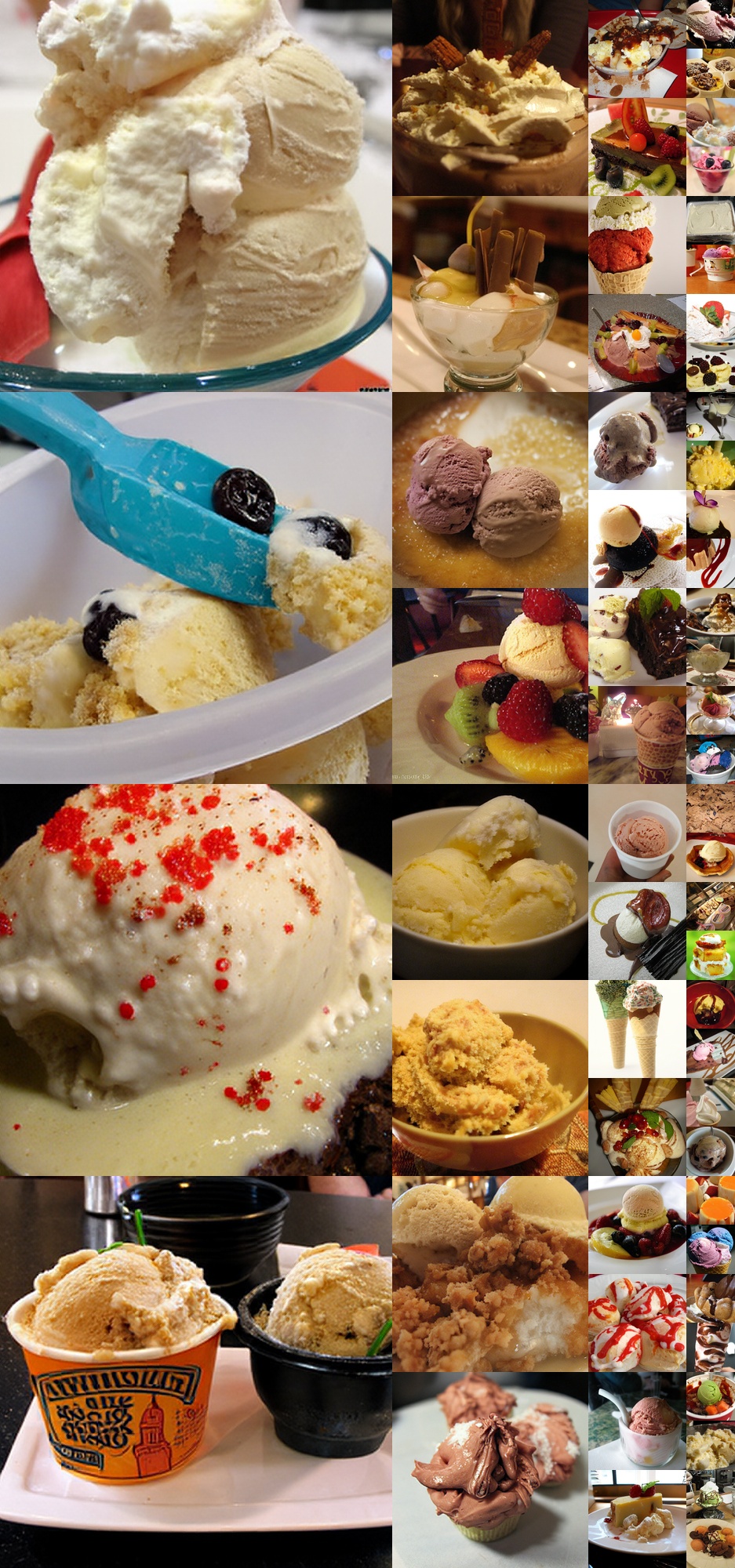}}
  \caption{
   \textbf{Uncurated $512 \times 512$ MixFlow-XL samples.}  \\
   AutoGuidance Scale = 1.5\\
   Class label = ``ice cream'' (928)}
 \label{fig:512_928}
\end{figure}

\begin{figure}[t]
  \centering
  \footnotesize
    \centering
{\includegraphics[width=1.0\linewidth]{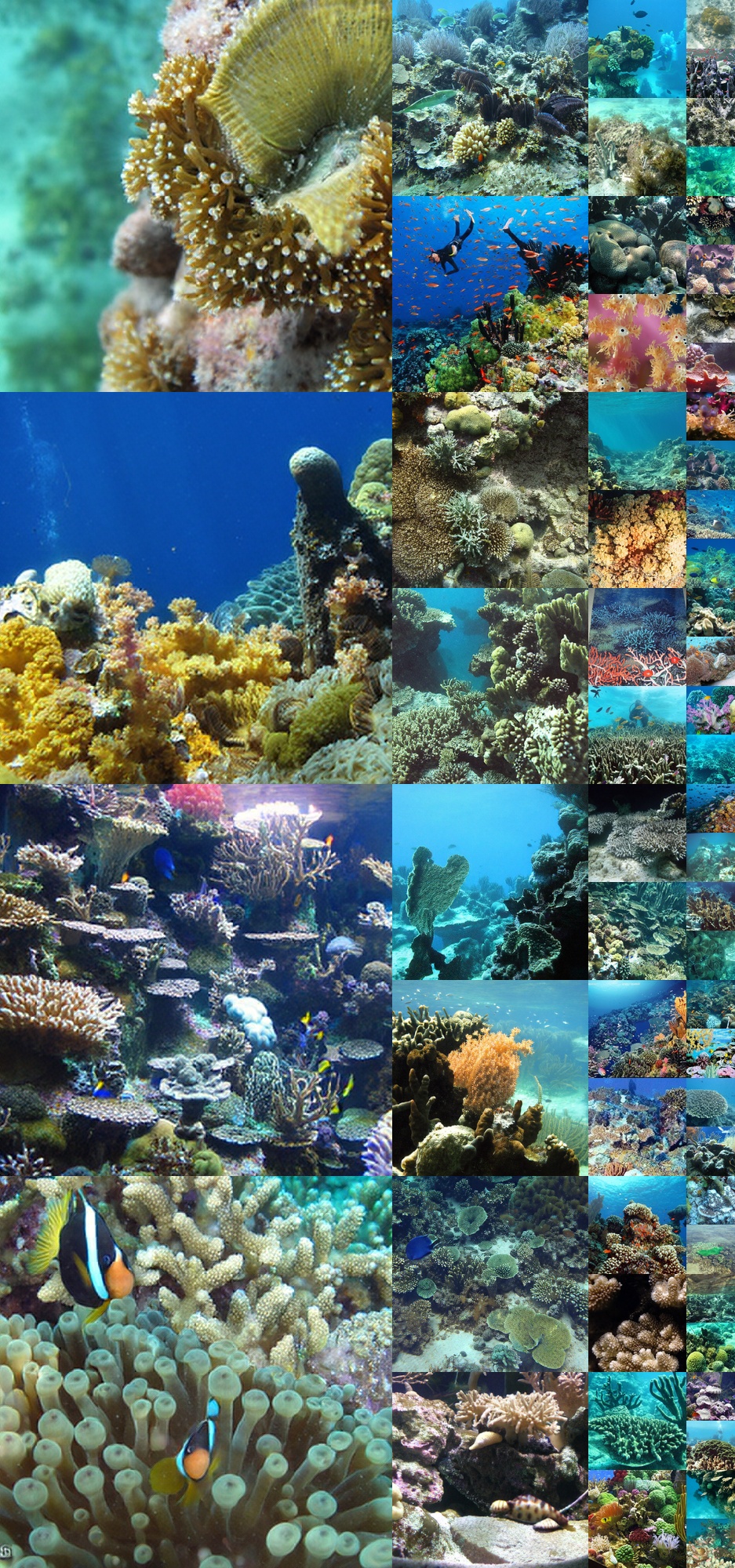}}
  \caption{
   \textbf{Uncurated $512 \times 512$ MixFlow-XL samples.}  \\
   AutoGuidance Scale = 1.5\\
   Class label = ``coral reef'' (973)}
 \label{fig:512_973}
\end{figure}

\noindent\textbf{Class-conditional generation on ImageNet $256 \times 256$ and $512 \times 512$.}
We present the visualization results for MixFlow + RAE class-conditional generation on ImageNet $256 \times 256$ and $512 \times 512$ 
in \Cref{fig:256_089,fig:256_207,fig:256_250,fig:256_270,fig:256_388,fig:256_417,fig:512_33,fig:512_88,fig:512_291,fig:512_387,fig:512_928,fig:512_973}.
The results demonstrate high semantic diversity and fine details.

\section{Example Code}
The MixFlow training implementation
is very simple.
For example,
only \textbf{6 lines} of code are modified
for the RAE implementation.
The modification example is shown in
Figure~\ref{fig:5linesmodified}.

\section{Extensions}
The initial results show that 
the MixFlow training algorithm also benefit
(1)
training the SiT model
from scratch
and training
the shortcut model~\cite{FransHLA25} for few-step sampling.

\clearpage

\begin{figure*}[t]
    \centering
    \footnotesize
    \lstset{
        language=Python,
        basicstyle=\ttfamily\footnotesize,
        breaklines=true,
        escapechar=|,  %
        keywordstyle=\bfseries\color{blue},
        commentstyle=\itshape\color{gray},
    }
    
    %
    
    
    \begin{minipage}[t]{0.48\linewidth}
        \textbf{(a) RAE: Uniform sampling
        for the training timestep}
\begin{lstlisting}
def sample(self, x1):
    """Sampling x0 & t based on shape of x1 (if needed)
      Args:
        x1 - data point; [batch, *dim]
    """
    
    x0 = th.randn_like(x1)
    dist_options = self.time_dist_type.split("_")
    t0, t1 = self.check_interval(self.train_eps, self.sample_eps)
    if dist_options[0] == "uniform":
        t = th.rand((x1.shape[0],)) * (t1 - t0) + t0
    # ...


    t = t.to(x1)
    t = self.time_dist_shift * t / (1 + (self.time_dist_shift - 1) * t)
    return t, x0, x1
\end{lstlisting}
    \end{minipage}
    \hfill
    \begin{minipage}[t]{0.48\linewidth}
        \textbf{(b) MixFlow: Beta sampling for the training timestep}
\begin{lstlisting}
def sample(self, x1, |\colorbox{hlt}{gamma=0.4}|):
    """Sampling x0 & t based on shape of x1 (if needed)
      Args:
        x1 - data point; [batch, *dim]
    """

    x0 = th.randn_like(x1)
    dist_options = self.time_dist_type.split("_")
    t0, t1 = self.check_interval(self.train_eps, self.sample_eps)
    if dist_options[0] == "uniform":
        t = th.rand((x1.shape[0],)) * (t1 - t0) + t0
    # ...
    # Sample t from Beta(2,1)
    |\colorbox{hlt}{t = 1 - th.sqrt(t) }|
    t = t.to(x1)
    # Sample mt from unshifted t
    |\colorbox{hlt}{mt = t + th.rand\_like(t) * gamma * (1 - t)}|
    # Apply shift to both t and mt
    t = time_dist_shift * t / (1 + (time_dist_shift - 1) * t)
    |\colorbox{hlt}{mt = time\_dist\_shift * mt / (1+(time\_dist\_shift-1) * mt)}|
    return t, |\colorbox{hlt}{mt,}| x0, x1
\end{lstlisting}
    \end{minipage}
    
    \vspace{1em}
    
    %
    
    \begin{minipage}[t]{0.48\linewidth}
        \textbf{(c) RAE: Standard training loss}
\begin{lstlisting}
def training_losses(self, model,  x1, model_kwargs=None):
    """Loss for training the score model
    Args:
    - model: backbone model; could be score, noise, or velocity
    - x1: datapoint
    - model_kwargs: additional arguments for the model
    """
    if model_kwargs == None:
        model_kwargs = {}
    
    t, x0, x1 = self.sample(x1)


    
    t, xt, ut = self.path_sampler.plan(t, x0, x1)
    model_output = model(xt, t, **model_kwargs)
    B, *_, C = xt.shape
    assert model_output.size() == (B, *xt.size()[1:-1], C)

    terms = {}
    terms['pred'] = model_output
    if self.model_type == ModelType.VELOCITY:
        terms['loss'] = mean_flat(((model_output - ut) ** 2))
    else: 
        # ...
            
    return terms
\end{lstlisting}
    \end{minipage}
    \hfill
    \begin{minipage}[t]{0.48\linewidth}
        \textbf{(d) MixFlow: Training loss with mixed slowed interpolations}
\begin{lstlisting}
def training_losses(self, model, x1, model_kwargs=None):
    """Loss for training the score model
    Args:
    - model: backbone model; could be score, noise, or velocity
    - x1: datapoint
    - model_kwargs: additional arguments for the model
    """
    if model_kwargs == None:
        model_kwargs = {}

    |\colorbox{hlt}{t, mt, x0, x1 = self.sample(x1)}|
    # Compute slowed interpolation
    |\colorbox{hlt}{\_, xt, ut = self.path\_sampler.plan(mt, x0, x1)}|
    model_output = model(xt, t, **model_kwargs)
    B, *_, C = xt.shape
    assert model_output.size() == (B, *xt.size()[1:-1], C)

    terms = {}
    terms['pred'] = model_output
    if self.model_type == ModelType.VELOCITY:
        terms['loss'] = mean_flat(((model_output - ut) ** 2))
    else:
        # ...

    return terms
\end{lstlisting}
    \end{minipage}

    \caption{\textbf{Only 6 lines of code are modified.} 
    for the RAE implementation.
    Two functions in 
\url{https://github.com/bytetriper/RAE/blob/main/src/stage2/transport/transport.py} are modified.
Left: RAE; Right: MixFlow + RAE.
    Note:
    in the RAE implementation ,
    $t=1$ corresponds to 
    the noise
    and $t=0$ corresponds to
    the clean data.
    \colorbox{hlt}{The lines}
    corresponds to the modified code.}
    \label{fig:5linesmodified}
\end{figure*}

\end{document}